\documentclass[authoryear,review,12pt]{elsarticle}

\usepackage[utf8]{inputenc}     % Codificación del archivo fuente en UTF8
\usepackage[T1]{fontenc}        % Incorpora fuente con acentos

\usepackage{graphicx}
\usepackage{amssymb}
\usepackage{caption}
\usepackage{subcaption}
\usepackage{adjustbox}

\usepackage[hyphens]{url}

\usepackage{lineno}

\usepackage{booktabs}			% Para incorporar tablas
\usepackage{multirow}
\usepackage{tabularx}
\usepackage{array}          % Requerido por tabularx
  \newcolumntype{C}{>{\centering\arraybackslash}X} 
\usepackage{hyphenat}
\usepackage[table,xcdraw]{xcolor}
\usepackage{float}

\biboptions{sort}

\journal{Computers and Electronics in Agriculture}

\begin{document}

\begin{frontmatter}


\title{A noise-robust acoustic method for recognizing foraging activities of grazing cattle}



\author[add1,add2]{Luciano~S.~Martinez-Rau\corref{cor1}\fnref{now}}
\cortext[cor1]{Corresponding author}
\fntext[now]{Present address: Department of Computer and Electrical Engineering, Mid Sweden University, Holmgatan, PO Box 85230 Sundsvall, Västernorrland, Sweden.}
\ead{luciano.martinezrau@miun.se}

\author[add1,add3]{José~O.~Chelotti}
\ead{jchelotti@sinc.unl.edu.ar}
\author[add1]{Mariano~Ferrero}
\ead{mferrero@sinc.unl.edu.ar}
\author[add4,add5]{Julio~R.~Galli}
\ead{jgalli@lidernet.com.ar}
\author[add6,add7]{Santiago~A.~Utsumi}
\ead{sutsumi@nmsu.edu}
\author[add5]{Alejandra~M.~Planisich}
\ead{aplanisich@gmail.com}
\author[add1,add8]{H.~Leonardo~Rufiner}
\ead{lrufiner@sinc.unl.edu.ar}
\author[add1]{Leonardo~L.~Giovanini}
\ead{lgiovanini@sinc.unl.edu.ar}

\address[add1]{Instituto de Investigación en Señales, Sistemas e Inteligencia Computacional, sinc(i), FICH-UNL/CONICET, Argentina}
\address[add2]{Department of Computer and Electrical Engineering, Mid Sweden University, Sundsvall, Sweden}
\address[add3]{TERRA Teaching and Research Center, University of Liège, Gembloux Agro-Bio Tech (ULiège-GxABT), 5030 Gembloux, Belgium}
\address[add4]{Instituto de Investigaciones en Ciencias Agrarias de Rosario, IICAR, UNR-CONICET, Argentina}
\address[add5]{Facultad de Ciencias Agrarias, Universidad Nacional de Rosario, Argentina}
\address[add6]{W.K. Kellogg Biological Station and Department of Animal Science, Michigan State University, United States}
\address[add7]{Department of Animal and Range Science, New Mexico State University, United States}
\address[add8]{Facultad de Ingeniería, Universidad Nacional de Entre Ríos, Argentina}

\begin{abstract}

Farmers must continuously improve their livestock production systems to remain competitive in the growing dairy market. Precision livestock farming technologies provide individualized monitoring of animals on commercial farms, optimizing livestock production. Continuous acoustic monitoring is a widely accepted sensing technique used to estimate the daily rumination and grazing time budget of free-ranging cattle. However, typical environmental and natural noises on pastures noticeably affect the performance limiting the practical application of current acoustic methods. In this study, we present the operating principle and generalization capability of an acoustic method called Noise-Robust Foraging Activity Recognizer (NRFAR). 
The proposed method determines foraging activity bouts by analyzing fixed-length segments of identified jaw movement events produced during grazing and rumination. The additive noise robustness of the NRFAR was evaluated for several signal-to-noise ratios using stationary Gaussian white noise and four different nonstationary natural noise sources. In noiseless conditions, NRFAR reached an average balanced accuracy of 86.4\%, outperforming two previous acoustic methods by more than 7.5\%. Furthermore, NRFAR performed better than previous acoustic methods in 77 of 80 evaluated noisy scenarios (53 cases with p<0.05). 
NRFAR has been shown to be effective in harsh free-ranging environments and could be used as a reliable solution to improve pasture management and monitor the health and welfare of dairy cows. The instrumentation and computational algorithms presented in this publication are protected by a pending patent application: AR P20220100910.\\
Web demo available at: \url{https://sinc.unl.edu.ar/web-demo/nrfar}
\end{abstract}


\end{frontmatter}
%\linenumbers

\section{Introduction}
\label{S:1}
The new and diverse precision livestock farming tools and applications significantly reduce farm labor~\citep{Lovarelli2020-fi,Tzanidakis2023-pt}. Precision livestock farming solutions allow individualized monitoring of animals to optimize herd management in most production systems~\citep{Michie2020-vy}. Monitoring the feeding behavior of livestock can provide valuable insights into animal welfare, including their nutrition, health, and performance~\citep{Banhazi2012-jh,GARCIA2020105826}. Changes in \textcolor{black}{grazing} patterns, periodicity, and duration can be used to inform the management of pasture allocation~\citep{connor_2015}\textcolor{black}{, while changes in} ruminant diets signal anxiety~\citep{BRISTOW2007626} or stress~\citep{abeni2017monitoring,SCHIRMANN20096052}, as well as an early indicator of diseases~\citep{Osei-Amponsah2020-wg,Paudyal2018-hl}, rumen health~\citep{Beauchemin2018-dy,Beauchemin1991-fd}, and the onset of parturition~\citep{Kovacs2017-yx,Pahl2014-ae} and estrus~\citep{Dolecheck2015-cw,Pahl2015-cx}.

Free-ranging cattle spend 40-80\% of their daily time budget on grazing and rumination activities~\citep{Kilgour2012-ny,phillips2002}. %Grazing involves searching, apprehending, chewing, and swallowing herbage and is defined by a nonpredefined sequence of ingestive jaw movement (JM) events associated with chews, bites, and composite chew-bites. 
%Grazing involves the process of searching, apprehending, chewing and swallowing herbage and is characterized by a sequence of ingestive jaw movement (JM) events associated with chews, bites, and composite chew-bites, without a fixed or predefined order.
A grazing bout involves the process of searching, apprehending, chewing, and swallowing herbage and is characterized by a sequence of ingestive jaw movement (JM) events associated with chews, bites, and composite chew-bites, without a fixed or predefined order.
A bite event involves the apprehending and severing of the herbage, a chew event involves crushing, grinding, and processing previously gathered herbage, and a chew-bite event occurs when herbage is apprehended, severed, and comminuted in the same JM~\citep{Ungar2006-fq}. 
Rumination is defined as the period of time during which an animal repeatedly regurgitates previously ingested food (cud) from its rumen, then chews the cud for 40-60~s, and re-swallows it. Rumination bouts begin with the first regurgitation and end with the last swallow~\citep{Beauchemin2018-dy,galli2020discriminative}.
Grazing and rumination involve JM-events taken at rates of 0.75-1.20 JM per second. Changes in the type and sequence of distinctive JM-events can be aggregated over time to determine the sequence and duration of foraging activities~\citep{Andriamandroso2016-eo}.

Feeding activity monitoring of cattle has primarily been approached through the use of noninvasive wearable sensors, including nose-band pressure, inertial measurement units, and microphone systems~\citep{Benos2021-nm,Stygar2021-fa}. Each sensing technique has its advantages and disadvantages depending on the environment and application. Current nose-band pressure sensors are combined with accelerometers to log data from JMs. Raw data are analyzed by software to determine foraging behaviors and provide specific information associated with them~\citep{Steinmetz2020-uu,WERNER2018138}. Human intervention is required to process the data recorded on a computer, making it not scalable for use on commercial farms~\citep{Riaboff2022-hd}. Sensors based on inertial measurement units are widely used to recognize multiple behaviors such as \textcolor{black}{grazing}, rumination, posture, and locomotion~\citep{Aquilani2022-do,CHAPA2020104262}. Although accelerometer-based sensors are typically used in indoor environments~\citep{Balasso2021-gp,Lovarelli2022-nz,Wu2022-if}, their use in outdoor environments has increased in the last years~\citep{ARABLOUEI2023100163,Cabezas,WANG2023107647}. One major drawback of inertial measurement units is their limited ability to estimate herbage intake in grazing~\citep{Wilkinson2020-tx}. Furthermore, the reliability of these sensors is highly dependent on their precise location, orientation, and secure clamping, making reproducing results difficult~\citep{Kamminga2018-mg,li2021data}. For this reason, acoustic sensors are preferred over former sensors for monitoring the foraging \textcolor{black}{and rumination} behaviors of cattle outdoors. Head-placed microphones allow to collect detailed information on ingestive behaviors~\citep{Laca1992-am}. Acoustic sensors are used to automatically recognize JM-events~\citep{Ferrero2023-hm,Li2021-vn}, estimate rumination and grazing bouts~\citep{vanrell2018regularity}, distinguish between plants and feedstuffs eaten~\citep{galli2020discriminative,Milone2012-me}, and estimate differences in dry matter intake~\citep{Galli2018-zt}.
%Despite progress, the scarcity of public datasets in this field makes the development of reliable acoustic methods challenging~\citep{Cockburn2020-lb}. As a result, there is room for improvement in the acoustic monitoring of free-grazing cattle.
\textcolor{black}{Despite progress, the evaluation of the generalization capabilities of motion- and acoustic-based monitoring solutions are limited due to the scarcity of public and standardized datasets~\citep{VANRELL2020105623,Martinez_Rau2023.10.18.562979}. As a result, there is room for improving the confidence in the acoustic monitoring of free-grazing cattle.}

In recent years, acoustic methods have been developed for recognizing foraging activities. \citet{vanrell2018regularity} developed a method based on the analysis of the autocorrelation of the acoustic signal for the recognition of foraging activities. 
This method operates offline because it requires storing several hours of acoustic recording to discover the regularity patterns in the signal. Offline operation introduces considerable delays in making inferences about foraging activities, which could be critical for the early detection of estrus~\citep{ALLRICH1993249,REITH20126416}.
The Bottom-Up Foraging Activity Recognizer (BUFAR) developed by~\citet{chelotti2020online} operates online, meaning that the incoming digital acoustic signal is processed as it is generated.
BUFAR analyzes 5-min segments of identified JM-events to determine grazing and rumination bouts, outperforming the method of~\citet{vanrell2018regularity} with significantly lower computational costs. 
More recently, \citet{CHELOTTI202369} proposed an online Jaw Movement segment-based Foraging Activity Recognizer (JMFAR) that outperforms BUFAR. This is achieved by analyzing information from JMs that have been detected but not yet classified, enabling the recognition of grazing and rumination bouts.
However, BUFAR and JMFAR exhibited an average confusion of approximately 10\% between grazing and rumination, indicating a need for improvement in the recognition of these activities.
Another significant drawback of these methods is their limited capability to recognize foraging activities in diverse operational conditions or in the presence of noise~\citep{CHELOTTI202369}. 
\textcolor{black}{To be useful in practical applications, acoustic foraging recognizers must work properly even under adverse noise and mismatch conditions, where variations in recording settings and environmental conditions are common}. Additionally, low computation demands make them feasible for embedding in an acoustic monitoring sensor~\citep{AQEELURREHMAN2014263}.
Motivated by this need, this paper describes in detail the operation\textcolor{black}{, noise robustness and generalization capability} of an alternative acoustic method for the recognition of grazing and rumination activities in free-range cattle. The proposed method involves a noise-robust methodology for detecting and classifying JM-events used to recognize foraging activities. In a recent proof-of-concept study, the implementation of the proposed method was assessed for real-time operation on a low-power microcontroller~\citep{Martinez_SAS2023}.
The main contributions of this work are: (\emph{i}) present an online acoustic method for estimating grazing and rumination bouts in cattle, characterized by a low computational cost. It classifies four classes of JM-events, which are analyzed in fixed-length segments to delimit activity bouts.
(\emph{ii}) The proposed method recognizes foraging activities in free-range environments under \textcolor{black}{different and }adverse acoustic conditions, using a robust JM event recognizer that is capable of identifying JM events under quiet and noisy operating conditions.
(\emph{iii}) Artificial noise sounds of different natures are used to simulate multiple adverse acoustic scenarios in controlled experiments~\citep{Skowronski2004}.

The rest of this paper is organized as follows: %Section~2 briefly describes a system for the recognition of foraging activities and analyzes the operation and limitations of BUFAR and JMFAR. Subsequently, the proposed algorithm is introduced. This section also outlines the acquisition of the datasets, the experimental setup, and the performance metric used to validate the algorithms. The comparative results for the proposed and former algorithms are shown in Section~3. Section~4 explains and discusses the results of this work. Finally, the main conclusions follow in Section~5.
\textcolor{black}{Section~2 briefly describes a system for recognizing foraging activities and analyzes the operation and limitations of BUFAR and JMFAR. 
Section~3 introduces the proposed algorithm. This section also outlines the acquisition of the datasets, the experimental setup, and the performance metric used to validate the algorithms. The comparative results for the proposed and former algorithms are shown in Section~4. Section~5 explains and discusses the results of this work. Finally, the main conclusions follow in Section~6.}

\section{\textcolor{black}{Current acoustic method analysis}}
In this section, a brief description and limitations of two current acoustic foraging activity recognizers, called BUFAR and JMFAR, are presented. Both methods follow the general structure of a typical pattern recognition system~\citep{Bishop2006,Martinez_Rau2020-lc} and can be represented by the common block diagram shown in Figure~\ref{fig_1}. A foraging activity recognizer can be analyzed at three temporal levels: bottom, middle, and top. These levels operate on the millisecond, second, and minute scales, respectively. A JM-event recognizer operates at both the bottom and middle levels to detect and classify different types of JM-events. 
The input digitized sound is conditioned, processed, and down-sampled using signal processing techniques to reduce the computational cost of the middle and top levels. The processed signals are used at the middle level for a JM detector based on adaptive thresholds.
When a JM is detected, a set of distinctive JM features are computed over a time window centered on the JM. Finally, a machine learning model uses the extracted set of JM features to classify the JM-event with a corresponding timestamp. 
The middle level provides JM information to the top level.
The top level buffers the JM information in nonoverlapping segments of 5-min duration. For each segment, a set of activity features is computed to serve as input to a classifier that determines the activity performed by the animal. Segments of 5-min duration store sufficient JM information data in the buffer to generate a confidence set of activity features, without significantly affecting the correct delimitation of foraging activity. 
Five-min duration agrees with the optimal segment duration value found in two previous studies~\citep{chelotti2020online, ROOK199789}.

As previously mentioned, the type and sequence of distinctive JM-events can be analyzed to recognize foraging activities. Inspired by this, the BUFAR method uses a real-time JM-event recognizer developed by~\citet{Chelotti2018-wg} to detect and classify JM-events into three different classes: chews, bites, and chew-bites. 
The JM information comprises the timestamps and classes of the JM-events (see the top level of Figure~\ref{fig_1}).
The JM information is analyzed in nonoverlapping 5-min segments. For each segment, a set of four statistical activity features is extracted, including (\emph{i}) the rate of JM-events, and the proportion of the JM-events corresponding to the classes (\emph{ii}) chew, (\emph{iii}) bite, and (\emph{iv}) chew-bite. These features are then used for a multilayer perceptron (MLP) classifier~\citep{Bishop2006} to determine the activities performed. However, inherent detection and classification errors of JM-events may cause misclassification of foraging activities. A more detailed description of BUFAR is provided by~\citet{chelotti2020online}.

The JMFAR method partially overcomes the limitation of BUFAR because it does not compute information about the JM-events classes.
Instead, JMFAR analyses nonoverlapping 5-min segments from the detected JM. 
The same JM-event recognizer used in BUFAR is also used in JMFAR to compute the JM information. 
JM information consists of the signal used to detect the JM, the timestamps of the detected JM, and the extracted set of JM features. JM information, analyzed in segments, is employed to compute a set of activity features. The set of twenty-one statistical, temporal, and spectral features serves as input to an MLP classifier that determines the corresponding activity performed. A more detailed description of JMFAR is provided by~\citet{CHELOTTI202369}.

\section{Material and Methods}
\subsection{Proposed foraging activity recognizer}
The high sensitivity to noise of the JM-event recognizer used in BUFAR and JMFAR could lead to the misclassification of foraging activities. When the input audio signal is contaminated by noise, the accurate detection of JM, the computation of JM features, and the classification of JM-events are significantly impacted~\citep{Martinez-Rau2022-ya}. As a result, the noise directly impacts the JM information and consequently affects the computation of the set of activity features, leading to possible misclassification of activity.
The activity recognition in quiet and noise conditions can be improved by using a better JM-event recognizer. This work proposes an online method called \emph{Noise-Robust Foraging Activity Recognizer} (NRFAR). 
NRFAR introduces the use of the Chew-Bite Energy Based Algorithm (CBEBA) for the recognition of JM-events in diverse acoustic environments~\citep{Martinez-Rau2022-ya}.
Similar to BUFAR, NRFAR analyses nonoverlapping segments of 5-min duration of recognized JM-events classes for the subsequent classification of foraging activities.

\begin{figure}[!htb]
   \centering
   \includegraphics[width=1\textwidth]{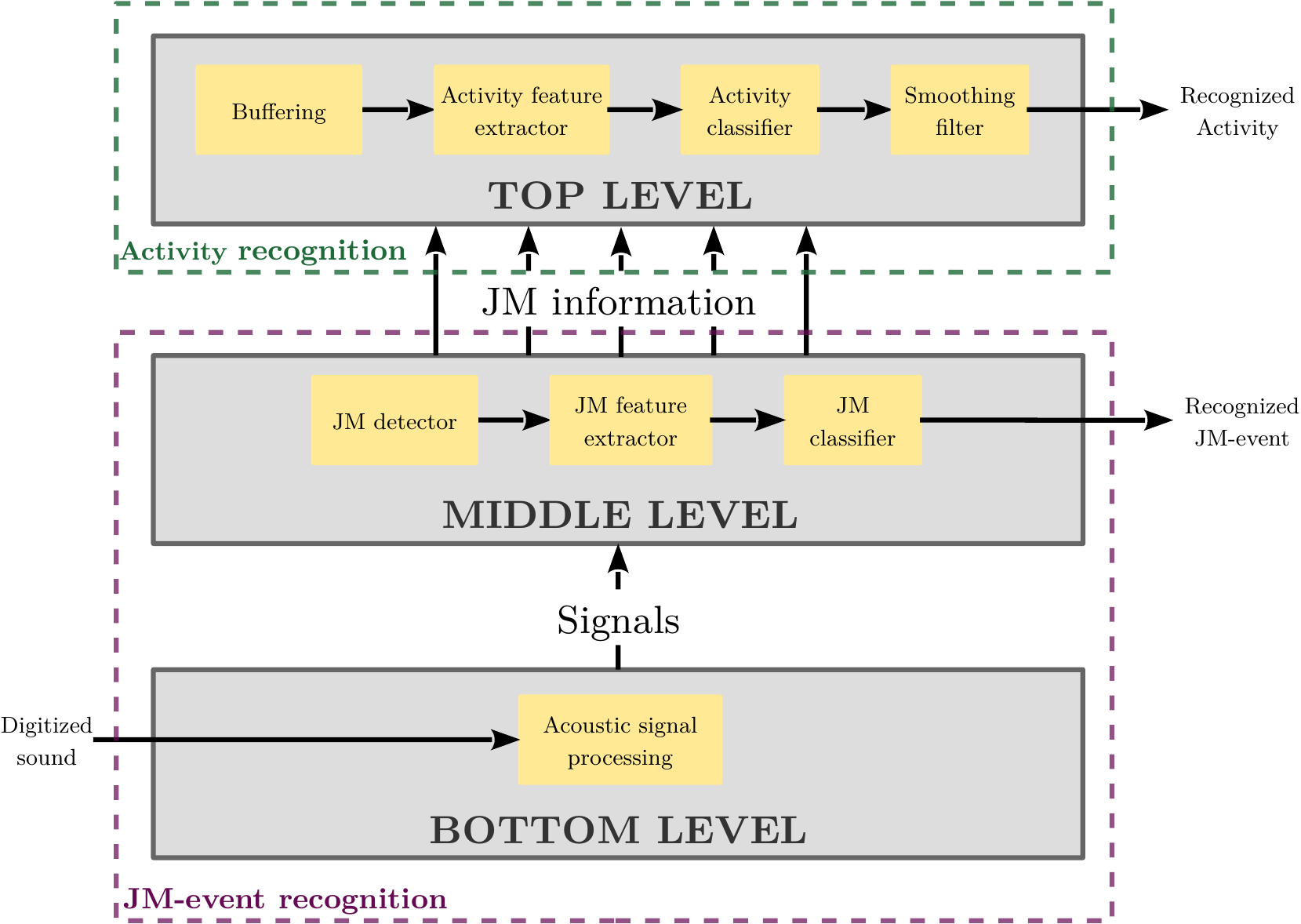}
   \caption{General block diagram of the BUFAR, JMFAR, and the proposed NRFAR methods divided into temporal scales. The JM information transferred to the top level is different in each method.}
   \label{fig_1}
\end{figure}

The CBEBA is a real-time pattern recognition method, able to distinguish individualized JM-events in terms of four different classes: \emph{rumination-chews}, \emph{grazing-chews}, \emph{bites}, and \emph{chew-bites}.
It outperforms previously published methods in both the detection and classification of JM-events in both noiseless and noisy environments. Briefly, the implementation of CBEBA can be divided into four successive stages (Figure~\ref{fig_1}):
\begin{itemize}
    \item 
    Signal processor: the digitized input audio signal undergoes a second-order Butterworth band-pass filter to isolate the JM frequency range. The filtered signal is then squared to obtain the instantaneous power signal. To reduce computation, the former signal is used to compute two additional down-sampled signals: a decimated envelope signal and an energy signal calculated by frames.
    \item 
    JM detector: the presence of a peak in the envelope signal above a time-varying threshold indicates the detection of a candidate JM-event. When this indication occurs, the energy signal is compared with another adaptive threshold to delimit the boundaries of the candidate JM-event.
    \textcolor{black}{The time-varying threshold considers short-timescale anatomical and behavioral characteristics of the animal, as well as, long-timescale variable feeding patterns. The adaptive threshold changes according to the background noise floor level on the acoustic signals.}
    \item 
    JM feature extractor: both delimited signals are used to extract a set of five robust JM features. These heuristic features are the duration, energy, symmetry of the envelope, zero-cross derivative of the envelope, and accumulated absolute value of the derivative of the envelope.
    \textcolor{black}{To avoid the detection of a false-positive JM-event, it is classified only if the duration and energy are in a predefined range.} 
    \item 
    %JM classifier: the computed set of JM features is used to decide whether the candidate JM-event should be classified or discarded. 
    JM classifier: A multilayer perceptron (MLP) classifier determines the class of the JM-event. The structure of the MLP classifier is 5-6-4 neurons in the input, hidden, and output layers.
    Furthermore, the adaptive thresholds are tuned based on the signal-to-noise ratio (SNR) estimated over the envelope and energy signals.
\end{itemize}

\textcolor{black}{A more detailed description of CBEBA is provided by \cite{Martinez-Rau2022-ya}}.

\begin{figure}[!htb]
   \centering
   \includegraphics[width=1\textwidth]{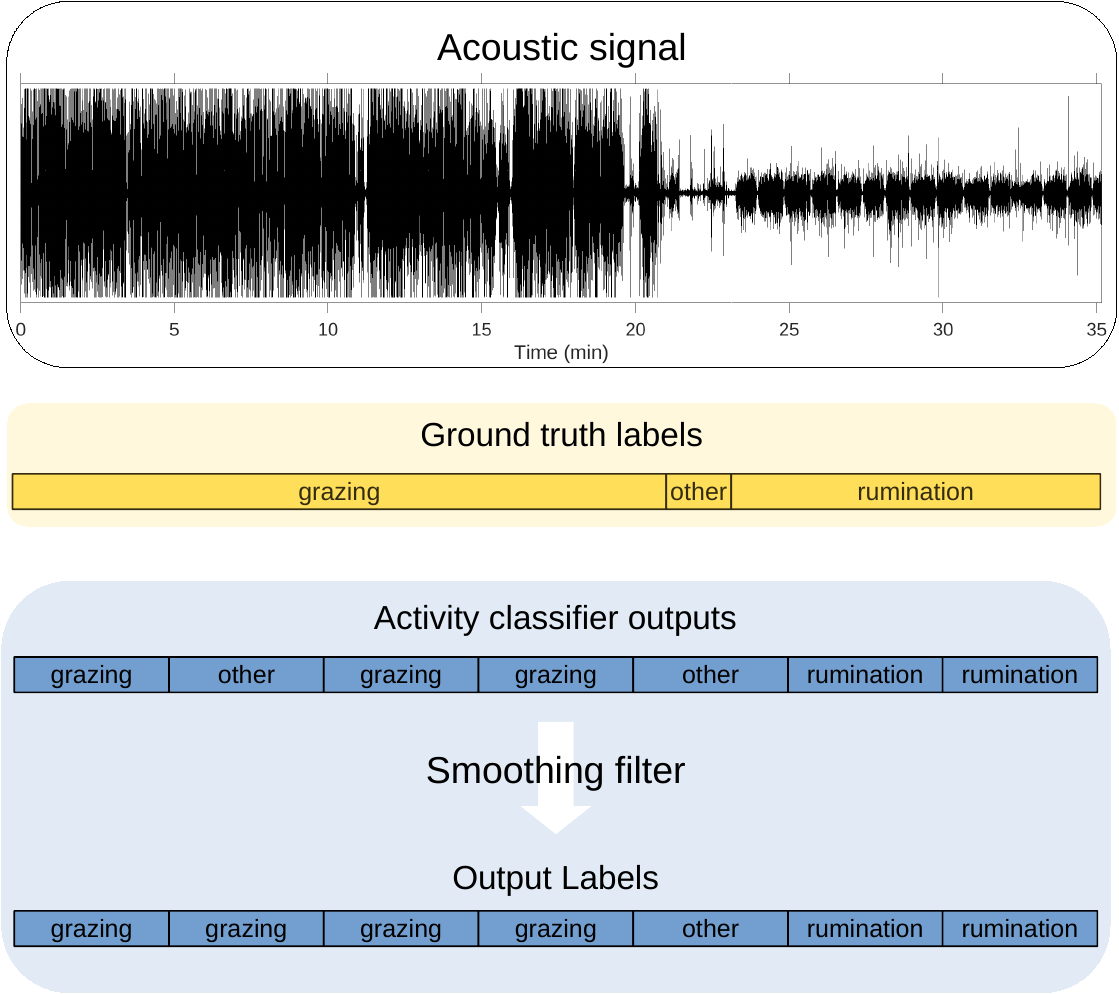}
   \caption{Example of recognized 5-min segments (blue color) compared to the ground truth reference labels (yellow color). The classified activity label assigned to every segment enters the smoothing filter to generate the output label of NRFAR.}
   \label{fig_2}
\end{figure}

The top level of the proposed NRFAR processes the JM information provided by the JM-event recognizer CBEBA in nonoverlapping 5-min segments to establish the corresponding foraging activity. The JM information is the recognized JM-events, along with their respective timestamps. Each segment of JM information is used to generate a set of five activity features: (\emph{i}) the rate of JM-events, and the proportion of the JM-events corresponding to the classes (\emph{ii}) rumination-chew, (\emph{iii}) grazing-chew, (\emph{iv}) bite, and (\emph{v}) chew-bite). 
The set of extracted activity features feeds an MLP activity classifier to label the foraging activity \textcolor{black}{in terms of \emph{grazing}, \emph{rumination} and \emph{other}}. The classified label outputs are smoothed using a third-order median filter to reduce the possible misclassifications of the recognized activity along consecutive segments. Figure~\ref{fig_2} shows an example of the proper operation of the smoothing filter.

%% FIGURA 1

%% FIGURA 2
\subsection{Datasets description}
%The fieldwork to obtain acoustic signals took place at the Michigan State University’s Pasture Dairy Research Center (W.K. Kellogg Biological Station, Hickory Corners, MI, USA) from July~31 to August~19, 2014. The procedures for animal handling, care, and use were revised and approved by the Institutional Animal Care and Use Committee of Michigan State University. 
\textcolor{black}{This study uses two datasets to evaluate the algorithms under matched and mismatched conditions.
The first one (referred to as DS1) is a public dataset collected at the Michigan State University’s Pasture Dairy Research Center (W.K. Kellogg Biological Station, Hickory Corners, MI, USA) from July~31 to August~19, 2014~\citep{Martinez_Rau2023.10.18.562979}. In this dataset,} the cows were handled using a pasture-based robotic milking system with unrestricted cow traffic, as described by~\citet{Watt2015-ek}. Cows were voluntarily milked~$3.0\pm1.0$ times per day using two Lely A3-Robotic milking units (Lely Industries NV, Maassluis, The Netherlands). Inside the dairy barn, the dairy cows were fed a grain-based concentrate. Cows had 24-h access to grazing paddocks with a predominance of either tall fescue (Lolium arundinacea), orchardgrass (Dactylis glomerata) and white clover (Trifolium repens), or perennial ryegrass (Lolium perenne) and white clover. From a herd of 146~lactating high-producing multiparous Holstein cows, 5~animals were selected to record acoustic signals and to monitor their foraging behavior in a noninvasive manner continuously. Specific information on grain-based concentrate, pasture on paddocks, and individualized characteristics of the 5~dairy cows are given in~\citet{Martinez_Rau2023.10.18.562979}.

Individualized 24-h of continuous acoustic recordings were obtained on 6~nonconsecutive days. The foraging behavior of the 5~dairy cows was recorded by 5~independent recording systems that were rotated daily, according to a 5~x~5 Latin-square design. This setup was allowed to verify differences in sound signals associated with a particular recording system, cow, or experimental day. The recording systems were randomly assigned to the cows on the first day. On the sixth day, the same order was used to reassign the recording systems to the cows. No prior training was considered necessary for the use of the recording systems before the start of the study. 

\begin{figure}[!htb]
   \centering
   \includegraphics[width=1\textwidth]{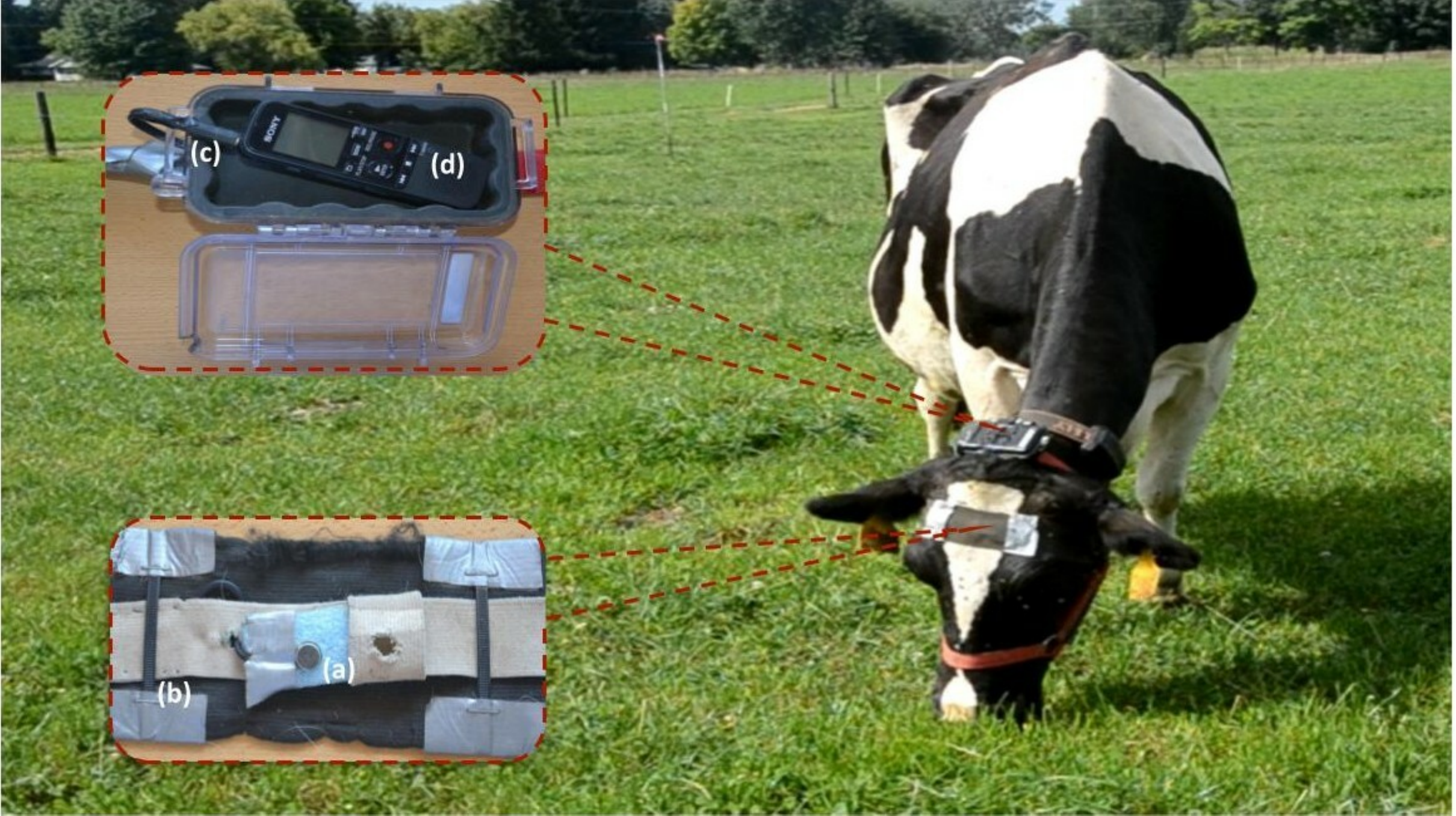}
   \caption{Recording system used to record the acoustic signals composed of microphones (a) that are covered by rubber foam and an elastic headband (b), which are wired and plugged (c) to a digital recorder placed inside a waterproof case (d) attached to a neck halter. \textcolor{black}{Figure extracted from \cite{Martinez_Rau2023.10.18.562979}}}
   \label{fig_3}
\end{figure}

Each recording system comprised two directional electret microphones connected to a digital recorder (Sony Digital ICD-PX312, Sony, San Diego, CA, USA). The digital recorder was protected in a weatherproof case (1015 Micron Case Series, Pelican Products, Torrance, CA, USA) and mounted on the top side of a halter neck strap (Figure~\ref{fig_3}). One microphone was positioned facing outwards in a noninvasive manner and pressed against the forehead of the cow to collect the sounds produced by the animal. The other microphone was placed facing inwards to capture the vibrations transmitted through the bones. The microphones kept the intended location using rubber foam and an elastic headband attached to the halter. This design prevents microphone movements, reduces wind noise, and protects microphones from friction and scratches~\citep{Milone2012-me}. The digital recorders saved the audio recordings in MP3 format~\citep{brandenburg1994iso} with a 16-bit resolution at a sampling rate of 44.1~kHz. 
Each channel of the stereo MP3 files corresponds to the microphone facing inwards and outwards. In this study, the stereo MP3 files were converted to mono WAV files, and only those mono WAV files corresponding to the microphones facing inwards were used because they provide a better sound quality of the foraging activities with less presence of external noise sounds.

%% FIGURA 3

\textcolor{black}{The second dataset (referred to as DS2) was collected at the Campo Experimental J.F. Villarino (Facultad de Ciencias Agrarias, Universidad Nacional de Rosario, Zavalla, Argentina) on August 1, 2022. The protocol used for the experiment has been evaluated and approved by the Committee on Ethical Use of Animals for Research of the Universidad Nacional de Rosario. This intensified pastoral-based dairy farm has a herd of 140-165 milking cows, with an individual production of 24-27 l of milk daily. Three 4-year-old lactating Holstein cows weighing 570-600 kg were selected for this experiment. The experimental cows were allowed to graze freely within a fully enclosed paddock measuring approximately 60 by 20 m, and they had continuous access to a watering trough. The paddock area was covered with naturalized perennial grasses (with a dominance of Cynodon sp., Lolium sp., and Festuca sp.). All cows were tamed and trained in the experimental routine before the experiment.}
\textcolor{black}{Each animal was equipped with an acquisition data device consisting of an external microphone (IP57 100 mm, -42 $\pm$ 3 dB, SNR 57 dB) plugged via a 3.5 mm jack to a Moto~G6 smartphone~\citep{motog6}. The smartphones were fixed inside plastic boxes secured to prevent unintended internal movements. As in DS1, microphones were located on the cow’s forehead and boxes were mounted to the top sides of halter neck straps (Figure~\ref{fig_3}). Audio recordings were stored in the Moto G6 using high-efficiency advanced audio coding \citep{bosi1997iso/iec} with a bit rate of 128 kbps and a sampling rate of 44.1 kHz, single channel (mono).}

\textcolor{black}{Each fieldwork} employed an experienced animal handler who had extensive knowledge of data collection on animal behavior. The handler observed the animals for blocks of approximately 5~min per h during daylight hours to ensure the proper placement and positioning of recording systems on the cows. The observations were conducted from a distance to minimize potential disruptions in animal behavior. The handler registered the observed foraging activities and other relevant parameters in a logbook. The ground truth identification of foraging activities was carried out by two experts with long experience in foraging behavior scouting and in the digital analysis of acoustic signals. An expert listened to the audio recordings to identify, delimit, and label the activities guided by the logbook. The results were double-inspected and verified by the other expert. Although the experts agreed on all label assignments, there were some small differences in the start or end times of certain labels. In these cases, the experts collaborated to reach a mutual agreement on the labels. Activity blocks were labeled as \emph{grazing}, \emph{rumination}, or \emph{other} (see Figure~\ref{fig_2}).

\begin{figure}[!htb]
   \centering
   \includegraphics[width=0.4\textwidth]{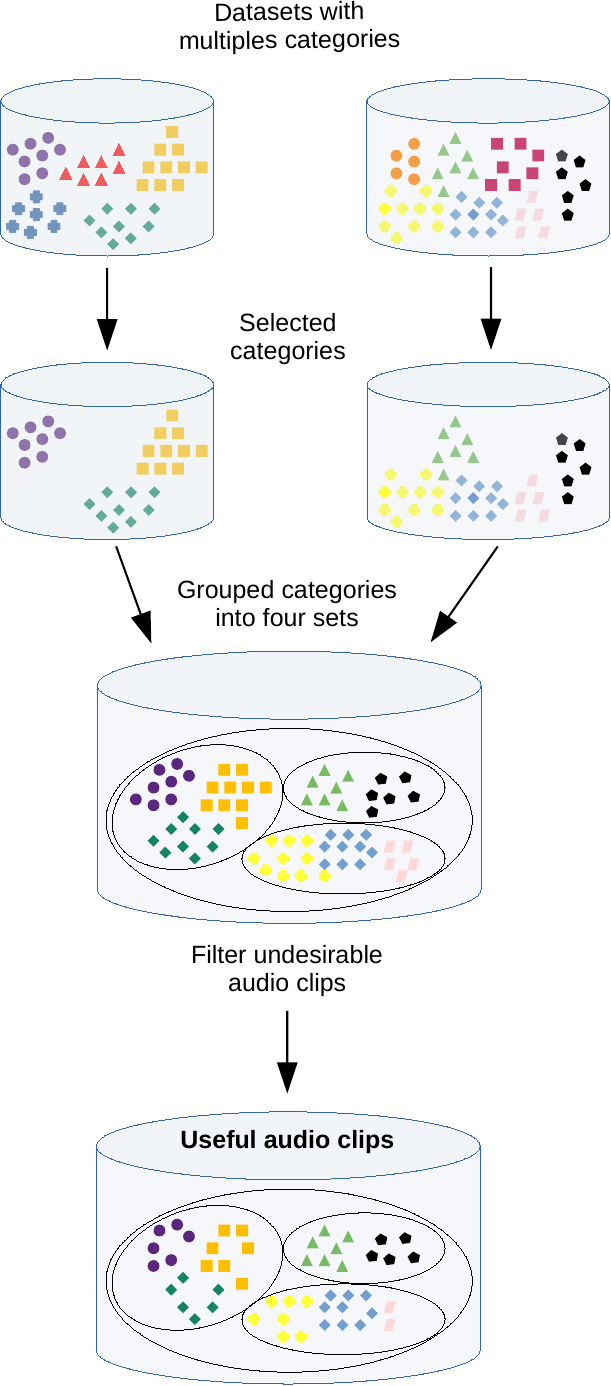}
   \caption{Top-down scheme for selecting useful audio clips.}
   \label{fig_4}
\end{figure}

\textcolor{black}{Additionally, this study uses audio clips from two open acoustic datasets to evaluate the algorithms under adverse conditions}. The selection process for the useful audio clips is shown in Figure~\ref{fig_4}. The first dataset is a labeled collection of 2000~environmental audio clips of 5~s duration, organized into 50~categories with 40~audio clips per category~\citep{Piczak2015-vx}. The second dataset is a multilabeled collection of 51,197~audio clips, with a mean duration of 7.6~s, unequally distributed into 200~categories~\citep{Fonseca2022-ta}. To represent environmental and natural noises commonly found in field pastures, the categories \emph{``aeroplane''}, \emph{``chirping birds''}, \emph{``cow''}, \emph{``crickets''}, \emph{``engine''}, \emph{``insects''}, \emph{``rain''}, \emph{``thunderstorm''}, and \emph{``wind''} from the first dataset and \emph{``aircraft''}, \emph{``animal''}, \emph{``bird vocalisation and birds call and bird song''}, \emph{``car passing by''}, \emph{``cowbell''}, \emph{``cricket''}, \emph{``engine''}, \emph{``fixed-wing aircraft and aeroplane''}, \emph{``frog''}, \emph{``insect''}, \emph{``livestock and farm animals and working animals''}, \emph{``rain''}, \emph{``raindrop''}, \emph{``thunder''}, and \emph{``wind''} from the second dataset were selected. These categories were grouped into four exclusive sets according to their nature as follows:

\begin{enumerate}
    \item 
    Animals = \{\emph{animal}, \emph{bird vocalisation and birds call and bird song}, \emph{chirping birds}, \emph{cow}, \emph{cowbell}, \emph{cricket}, \emph{crickets}, \emph{frog}, \emph{insect}, \emph{insects}, \emph{livestock and farm animals and working animals}\}
    \item
    Vehicles = \{\emph{aeroplane}, \emph{aircraft}, \emph{car passing by}, \emph{engine}, \emph{fixed-wing aircraft and aeroplane}\}
    \item
    Weather = \{\emph{rain}, \emph{raindrop}, \emph{thunder}, \emph{thunderstorm}, \emph{wind}\}
    \item
    Mixture = \{\emph{Animals}, \emph{Vehicles}, \emph{Weather}\}
\end{enumerate}

The audio clips of the sets were listened to by the experts, and those that did not correspond with possible field pasture conditions were discarded. Overall, 3042~useful audio clips lasting 13.1~h were identified. For reproducibility, a list of selected audio clips is available as Supplementary Material.

%% FIGURA 4

\subsection{Numerical experiments setup}
\subsubsection{\textcolor{black}{Experiment 1: performance evaluation under matched conditions}}
\textcolor{black}{
%In this study, three different experiments have been conducted. 
In the initial experiment, the NRFAR performance was evaluated using DS1. This experiment assessed NRFAR effectiveness under consistent conditions, including the same animals, recording devices, and field conditions.} NRFAR was coded, trained, and tested in Matlab R2019b (MathWorks, Natick, MA, USA), following a stratified 5-fold cross-validation scheme.
%In this study, a set of 349.4~h of outdoor audio recordings, composed of 50.5\% \emph{grazing}, 34.9\% \emph{rumination}, and 14.6\% of \emph{other} activities was used. 
\textcolor{black}{A set of 349.4~h of outdoor audio recordings of DS1, composed of 50.5\% \emph{grazing}, 34.9\% \emph{rumination}, and 14.6\% of \emph{other} activities was used.}
The imbalanced distribution of classes is consistent with typical cattle behavior~\citep{Kilgour2012-ny}. Therefore, the test data were not balanced by class. From all available training data in each fold, 30\% of the majority class (\emph{grazing}) was randomly undersampled and 100\% of the minority class (\emph{other}) was synthetically oversampled~\citep{he2008adasyn}, to generate a balanced dataset for training (35.6\% \emph{grazing}, 35.1\% \emph{rumination}, and 29.3\% of \emph{other} activities). The activity classifier is an MLP neural network formed by five input neurons (number of input features), one hidden layer, and three output neurons (number of output labels corresponding to the activity class). The activation functions used by the hidden and output layers are the hyperbolic tangent sigmoid and softmax transfer functions, respectively. During the MLP training phase, the scaled conjugate gradient backpropagation algorithm was used to find the optimal weight and bias of the network and optimize the MLP classifier's hyperparameters. The two hyper-parameters' learning rate and number of neurons in the hidden layer were fitted using a grid-search method. The learning rate was evaluated at values of 0.1, 0.01, 0.001, and 0.0001, whereas the number of neurons was evaluated within a range of~4 to~10.

\subsubsection{\textcolor{black}{Experiment 2: Generalization capability under clean mismatched conditions}}
\textcolor{black}{The NRFAR generalization capability was evaluated by processing acoustic signals from different animals located in another field and recorded with different devices. NRFAR was trained on DS1 and tested on DS2. The training set was balanced using the same under- and over-sampling techniques applied in the first experiment.
DS2 is composed of 13.2 h of audio recordings, corresponding to 51.8\% \emph{grazing}, 24.6\% \emph{rumination}, and 23.6\% of \emph{other} activities.}

\subsubsection{\textcolor{black}{Experiment 3: Noise robustness evaluation}}
External noise may reduce the operability of acoustic foraging activity recognizers operating under free-range conditions. The particular properties of these noise sources, including their finite duration and limited bandwidth, make them difficult to distinguish and quantify in the context of this study, which analyzed almost 350~h of audio recordings. Although audio recordings captured in \textcolor{black}{DS1} %the fieldwork 
might occasionally contain some noise, the signals were assumed to be free of noise; that is, they had an infinite SNR. \textcolor{black}{In this experiment, the }%The 
robustness of the NRFAR to noise was evaluated in five trials for various levels of contamination with noise and measured in terms of the SNR in a range from 20~to~-15 dB in steps of 5~dB. 
In each trial, \textcolor{black}{NRFAR was trained in the same way as in the first experiment but} a different noise source was artificially added to the audio recording \textcolor{black}{of DS1} used for testing and then normalized. A stationary Gaussian white noise source was used in a trial, which is one of the most accepted methods for testing the algorithm noise robustness~\citep{saez2016evaluating}. White noise is an \emph{``infinite''} bandwidth signal with constant power spectral density across all frequencies. Furthermore, the previously mentioned set of audio clips (\emph{Animals}, \emph{Vehicles}, \emph{Weather}, and \emph{Mixture}) was used in four trials to represent nonstationary environmental and natural noises present on the pasture. In each trial, the audio clips were randomly selected without replacement and concatenated to represent the artificial noise source that was used to contaminate the original audio recordings. Some examples of waveforms and spectrograms at several SNRs produced during grazing and rumination are shown in the Supplementary Material.

\subsection{\textcolor{black}{Metrics}}
State-of-the-art BUFAR and JMFAR methods were evaluated under the same conditions as NRFAR to establish a comparison between different methods.
Each audio recording has an associated ground-truth text file, specifying the start and end of the bouts, and the corresponding activity labels. The activity bouts, which last from several minutes to hours, were divided into nonoverlapping 1-s frames, following the approach described by~\citet{CHELOTTI202369}. This allowed a high-resolution activity recognition analysis to evaluate the performance of the methods.
This action was performed on both the algorithm output and the ground truth for a direct comparison. 
%In total, 1,257,759~frames were generated from the 349.4~h of audio recordings. This total number corresponds to 635,291, 439,262, and 183,206~frames for \emph{grazing}, \emph{rumination}, and \emph{other} activities, respectively. 
\textcolor{black}{In total, 1,257,759~frames and 47,606~frames were generated from the 349.4~h and 13.2~h of audio recordings of DS1 and DS2, respectively.}
For each audio signal, the balanced accuracy metric was calculated using the scikit-learn~1.2.2 library in Python\footnote{\url{https://scikit-learn.org/stable/modules/generated/sklearn.metrics.balanced_accuracy_score.html}}~\citep{scikit-learn}. This metric provides a good indicator of the performance of multiclass imbalance problems~\citep{mosley2013balanced}.

\section{Results}
\subsection{\textcolor{black}{Experiment 1}}
\begin{figure}[!htb]
   \centering
   \includegraphics[width=1\textwidth]{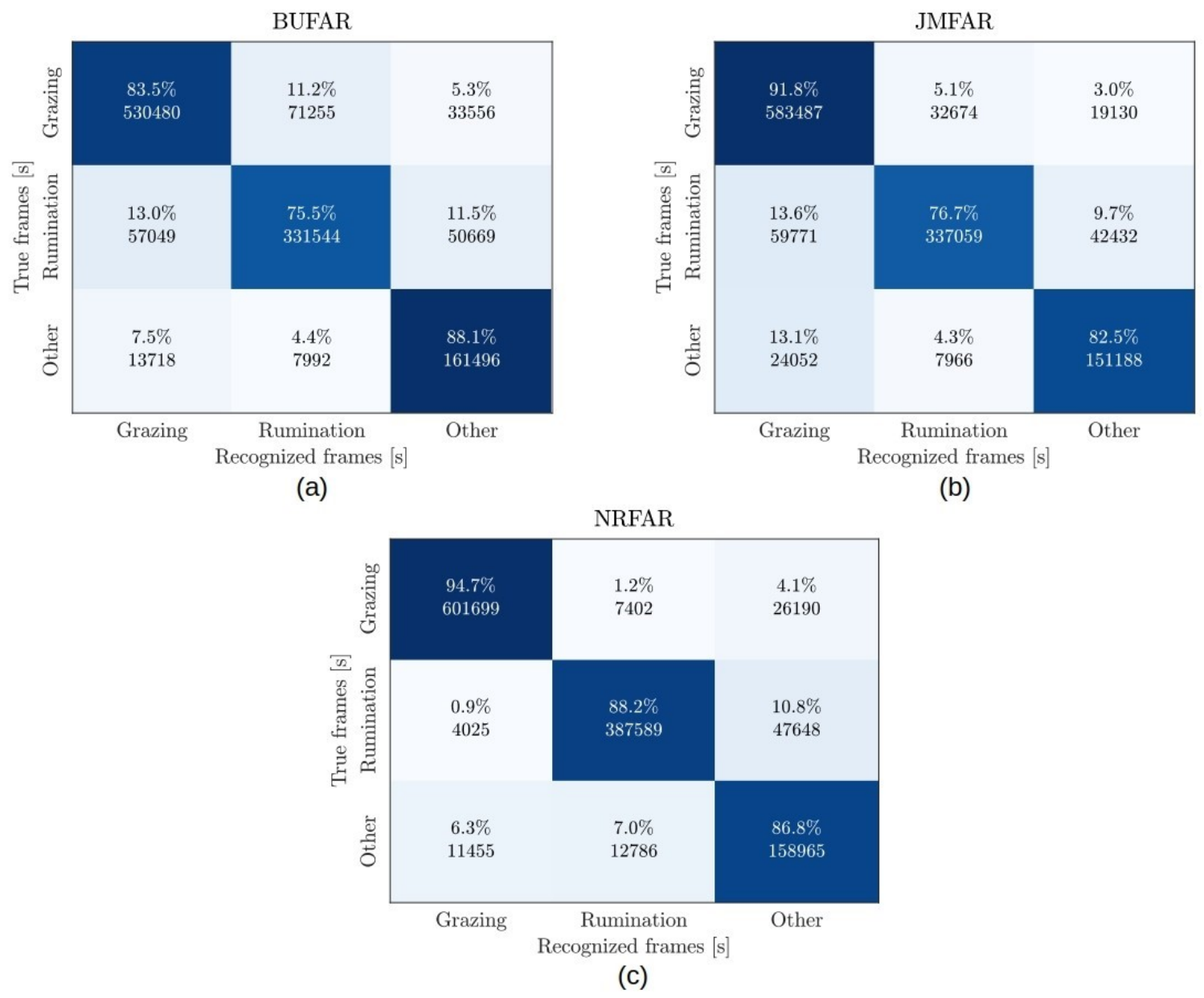}
   \caption{Confusion matrices for different foraging activities for the (a) BUFAR, (b) JMFAR, and (c) NRFAR methods \textcolor{black}{when evaluating on DS1}.}
   \label{fig_5}
\end{figure}

\textcolor{black}{The recognition performance of the different methods under matched conditions (i.e. trained and tested on DS1) reveals that }NRFAR properly classifies $\geq88.2\%$ of the frames into \emph{grazing} or \emph{rumination} classes, thus showing a significant improvement compared with the average of 79.5\% for BUFAR and 84.3\% for JMFAR (Figure~\ref{fig_5}). BUFAR exhibits the lowest recognition rate for the activities of interest but the highest recognition for \emph{other} activities (88.1\%). Moreover, confusion between \emph{grazing} and \emph{rumination} is lower for NRFAR ($\leq1.2\%$), than for BUFAR ($\geq11.2\%$) and JMFAR ($\geq5.1\%$). 

The computational cost of NRFAR, expressed in terms of operations per second (ops/s), is 13.4\% higher than that of BUFAR (43,060 ops/s vs. 37,966 ops/s) and 14.6\% lower than that of JMFAR (43,060~ops/s vs. 50,445~ops/s), with marginal variations presented between them. A detailed analysis and assumption of the operations involved are available in~\ref{Appendix_A}.

%% FIGURA 5

\subsection{\textcolor{black}{Experiment 2}}
\begin{figure}[!htb]
   \centering
   \includegraphics[width=1\textwidth]{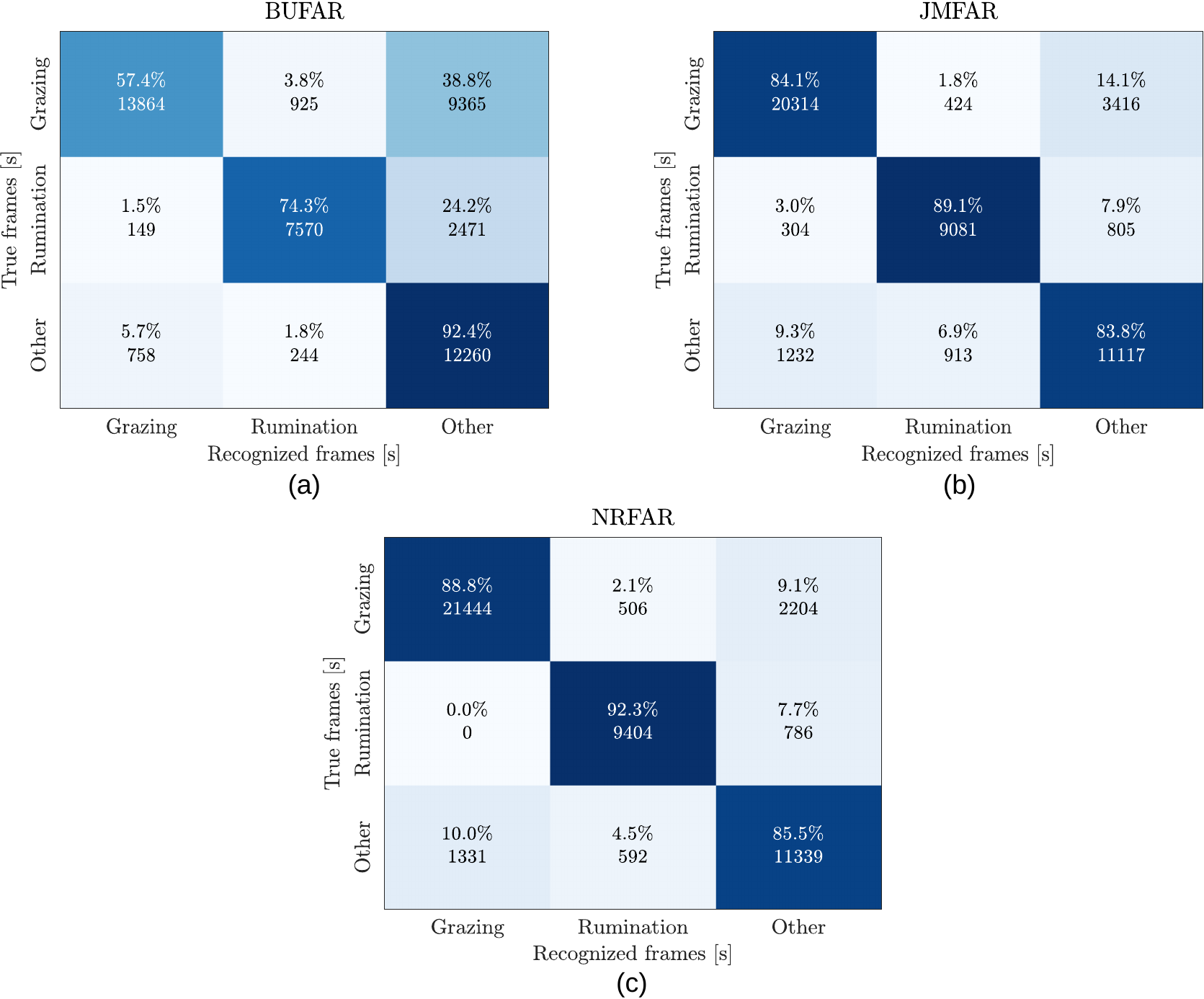}
   \caption{\textcolor{black}{Confusion matrices for different foraging activities for the (a) BUFAR, (b) JMFAR, and (c) NRFAR methods when evaluating on DS2}.}
   \label{fig_5_new}
\end{figure}

\textcolor{black}{The generalization capability of the different methods to recognize foraging activities is evaluated in the independent DS2 dataset. Figure~\ref{fig_5_new} shows the confusion matrices for the three methods. Qualitative previous results on DS1 are extended to those on DS2: NRFAR achieves a higher recognition rate for both \emph{grazing} and \emph{rumination} classes than JMFAR and BUFAR, with lower confusion between these classes.}

\textcolor{black}{The comparison of each method's performance in each dataset shows that NRFAR presents similar average balanced accuracies, being 86.4\% in DS1 and 87.4\% in DS2. Comparing Figure~\ref{fig_5_new}c versus Figure~\ref{fig_5}c, \emph{grazing} is 5.9\% higher in DS1 than in DS2, while \emph{rumination} is 4.1\% lower.
On the other hand, JMFAR exhibits a 7.7\% higher classification of \emph{grazing} but 12.7\% lower classification of \emph{rumination} in DS1 than in DS2 (Figure~\ref{fig_5_new}b versus Figure~\ref{fig_5}b).
The classification of \emph{other} activity is similar in DS1 and DS2 for both NRFAR and JMFAR.
BUFAR presents a similar capability for classifying \emph{rumination} in DS1 and DS2. However, the classification of \emph{grazing} decreases 26.1\% from DS1 to DS2 (Figure~\ref{fig_5_new}a versus Figure~\ref{fig_5}a).}

%% NUEVA FIGURA 5

\subsection{\textcolor{black}{Experiment 3}}
\begin{figure}[!htb]
   \centering
   \includegraphics[width=0.7\textwidth]{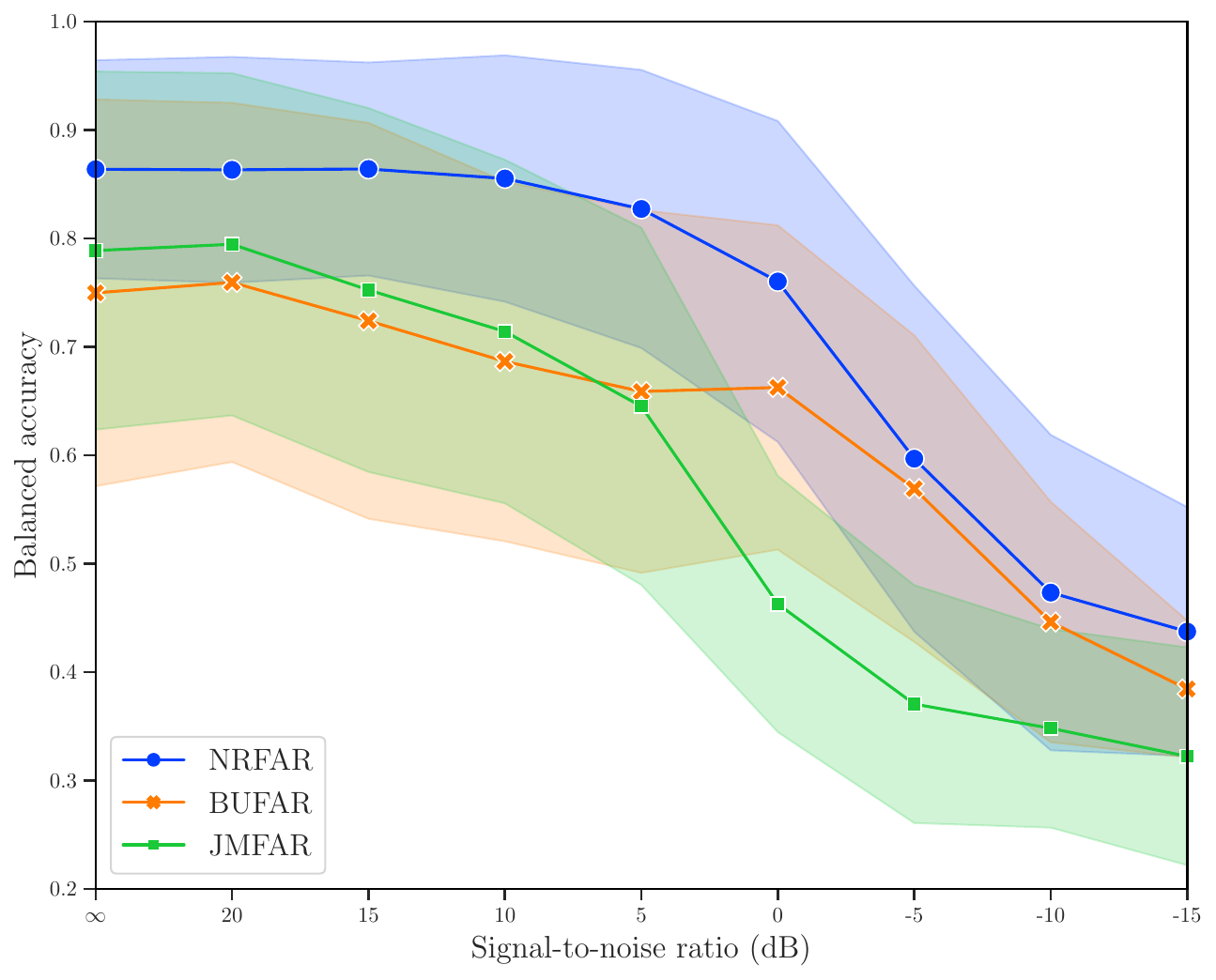}
   \caption{Performance rates (average~$\pm$~standard deviation) for the NRFAR, BUFAR, and JMFAR methods using additive Gaussian white noise at several SNR levels.}
   \label{fig_6}
\end{figure}

The robustness to adverse conditions of the NRFAR method is evaluated and compared against the BUFAR and JMFAR methods using different noise sources at multiple SNR levels. Gaussian white noise is added to the audio signals \textcolor{black}{of DS1} in appropriate proportions, to achieve the desired SNR. Figure~\ref{fig_6} shows the balanced accuracy, averaged over the audio signals, obtained with each method under different SNR conditions. NRFAR outperforms JMFAR and BUFAR in all cases (p~<~0.05; Wilcoxon signed-rank test \textcolor{black}{computed over the balanced accuracy of each signal}~\citep{Wilcoxon1945-zq}). The overall performance (average~$\pm$~standard deviation) of NRFAR remains approximately constant, ranging from $0.86~\pm~0.10$ to $0.83~\pm~0.13$ for SNR$~\geq 5$~dB.
Furthermore, the performance of JMFAR is higher (ranging from $0.79~\pm~0.16$ to $0.71~\pm~0.16$) than that of BUFAR (ranging from $0.76~\pm~0.17$ to $0.69~\pm~0.17$) under low noise conditions (SNR$~\geq 10$~dB). For moderate and high noise conditions (SNR$~\leq 5$~dB), BUFAR (ranging from $0.66~\pm~0.17$ to $0.39~\pm~0.06$) outperformed JMFAR (ranging from $0.65~\pm~0.16$ to $0.32~\pm~0.10$).

%% FIGURA 6

\begin{figure}[!htb]
   \centering
   \includegraphics[width=1\textwidth]{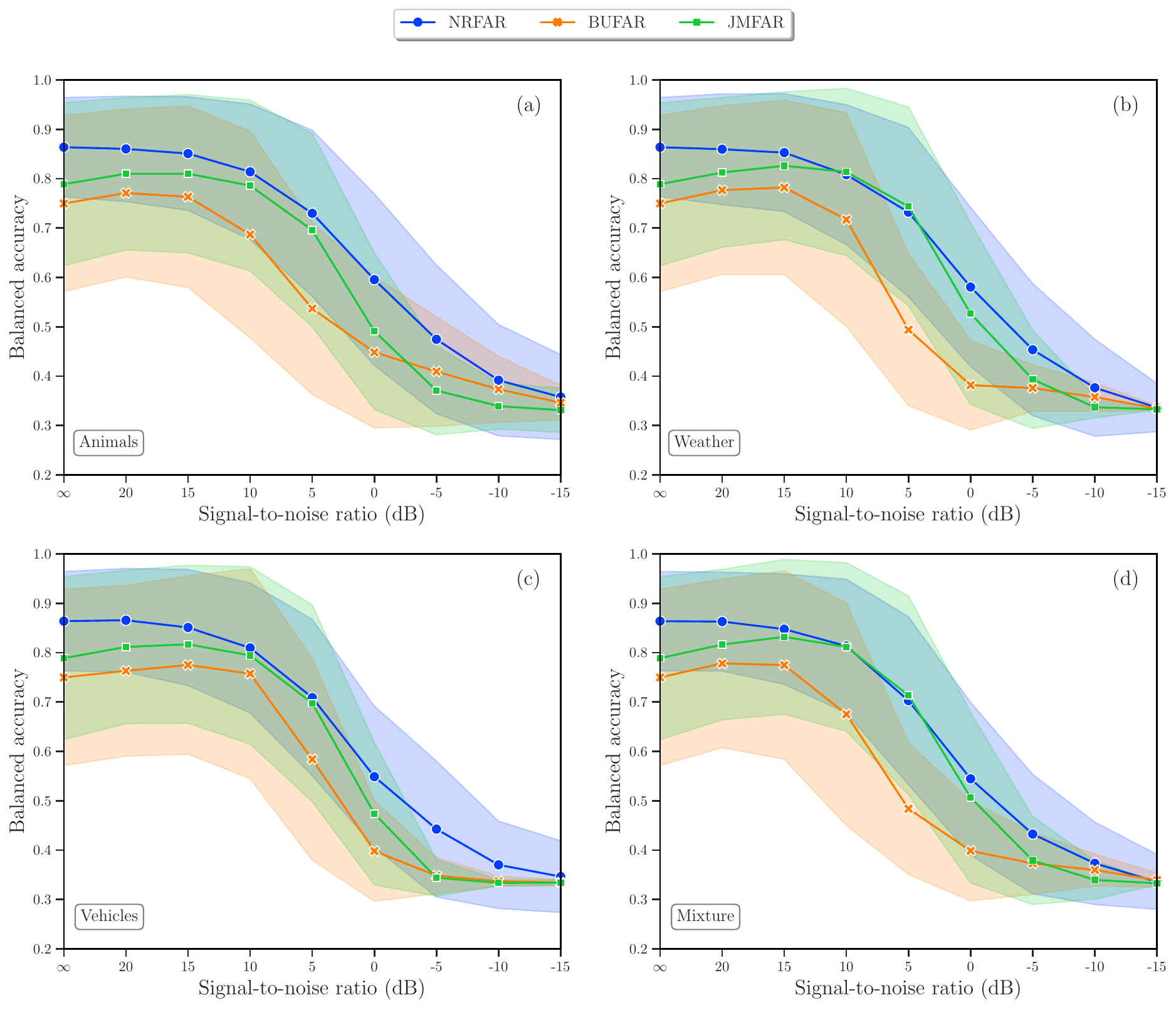}
   \caption{Performance rates (average~$\pm$~standard deviation) for the NRFAR, BUFAR, and JMFAR methods using noises commonly present on pasture at several SNR levels.}
   \label{fig_7}
\end{figure}

In a more challenging and realistic scenario, the original audio signals \textcolor{black}{of DS1} are mixed with a nonstationary noise source in four independent trials. The noise source contains exclusively sounds of animals, vehicles, weather, or a mixture of these sounds.
The balanced accuracy metrics reported by the methods using the four noise sources are shown in Figure~\ref{fig_7}. The performance of NRFAR decreases as the SNR decreases. However, the performance of BUFAR and JMFAR increases in general for SNR between 20~dB and 10~dB. In general, NRFAR outperforms BUFAR and JMFAR, particularly for SNR$~\geq15$~dB and for SNR$~\leq~0$~dB. 
NRFAR has a higher balanced accuracy than BUFAR in the 32 evaluated cases (p~<~0.05 in 25 cases).
Additionally, NRFAR outperforms JMFAR for SNR$~\geq~20$~dB and SNR$~\leq~0$~dB (p~<~0.05 in 14 of 16 cases).
The results of comparing NRFAR with JMFAR for SNR between 15~dB and 5~dB are not always statistically significant, although NRFAR presents higher performances than JMFAR in most cases (Figure~\ref{fig_7}).
On the other hand, JMFAR presents higher average balanced accuracy than BUFAR for SNR$~\geq~0$~dB for the four noise sources, particularly for $10~\geq~$SNR$~\geq~0$~dB (with p~<~0.05 in 19 of 20 cases).
Reported statistical significance test values obtained in the experiments are available in~\ref{Appendix_B}.

%% FIGURA 7

\begin{figure}[!htb]
   \centering
   \includegraphics[height=0.8\textheight]{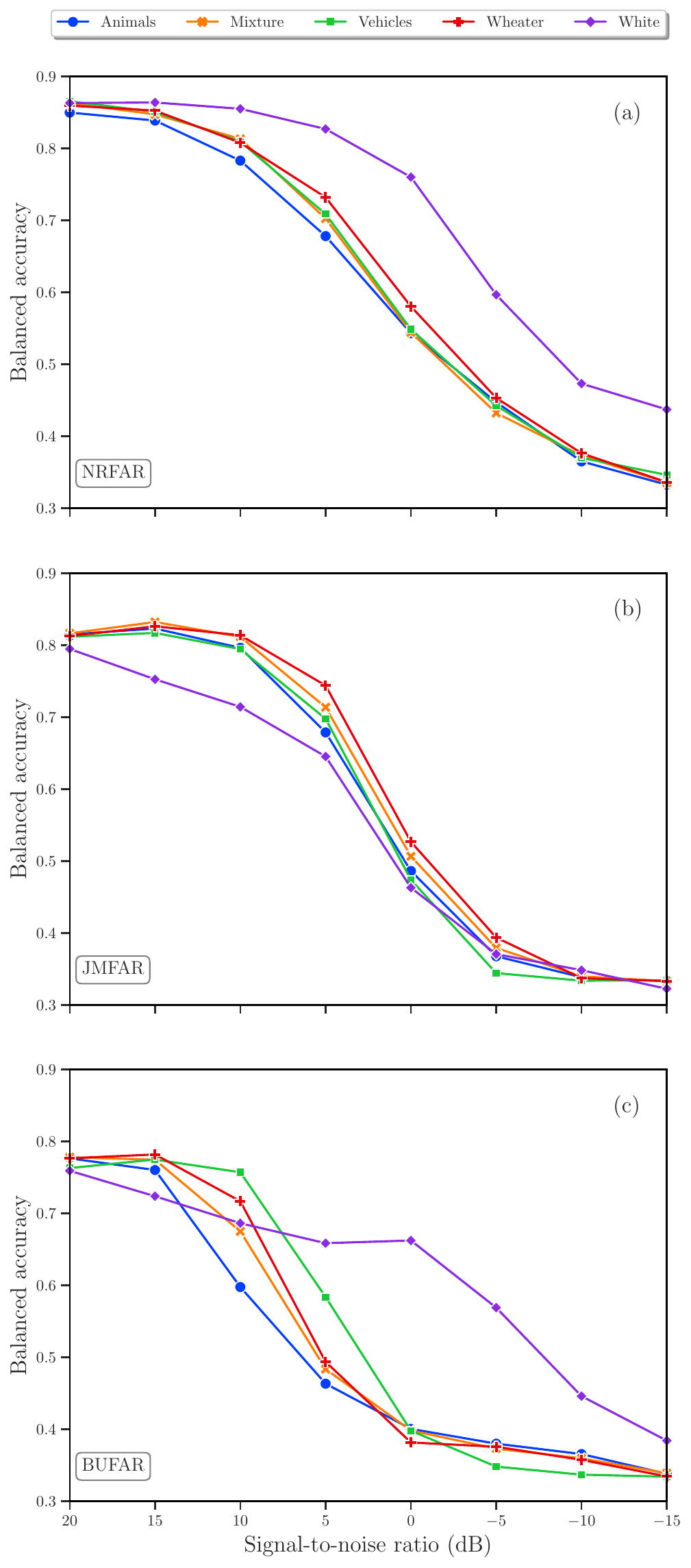}
   \caption{Variation of the performance metric across different noise sources for (a) NRFAR, (b) JMFAR, and (c) BUFAR. Marked points are the balanced accuracy, averaged over signals at a particular SNR level.
}
   \label{fig_8}
\end{figure}

The previously reported results have been rearranged to provide a different interpretation. Figure~\ref{fig_8} shows the performance degradation of the NRFAR, JMFAR, and BUFAR methods for the different noise sources. 
In Fig~\ref{fig_8}.a, the average balanced accuracy of NRFAR ranges from [0.86~-~0.85] for SNR~=~20~dB to [0.44~-~0.33] for~-15~dB. 
NRFAR reaches higher performance when Gaussian white noise is used. 
For a particular SNR value, NRFAR performs similarly between the noise sources representing more realistic acoustic pasture conditions. 
This is also true for JMFAR (Figure~\ref{fig_8}.b) but not for BUFAR (Figure~\ref{fig_8}.c). 

By comparing stationary and nonstationary noise sources, BUFAR and NRFAR exhibit higher performance when Gaussian white noise is added to the audio signals in moderate and high levels (SNR$~\leq 5$~dB). However, for low noise conditions, the recognition performance of JMFAR is more affected when Gaussian white noise is used.

%% FIGURA 8

\section{\textcolor{black}{Discussion}}
Accurately classifying the most important ruminant foraging behavior provides useful information to monitor their welfare and health, and to gain insight into their pasture dry matter intake and utilization~\citep{liakos2018machine}. This is typically achieved using accelerometers, pressure, or acoustic sensors. 
Commercial nose-band pressure sensors require handlers to analyze raw data recorded on a computer, which are not suitable for use in big rodeos~\citep{Riaboff2022-hd}. Ensuring the proper location, orientation, and attachment of accelerometer sensors mounted on a collar can become a laborious task for handlers to prevent their motion.
Meeting these requirements is even more challenging under free-ranging conditions. Therefore, acoustic sensors are preferable for practical use under such conditions~\citep{shen2020automatic}. Existing state-of-the-art acoustic methods for estimating the foraging activities of cattle, called BUFAR and JMFAR, are based on the analysis of fixed-length segments of sound signals. However, the misclassification of foraging activities remains a challenge.
This study proposes an improved online acoustic foraging activity recognizer (NRFAR) that analyzes identified JM-event classes in nonoverlapping segments of 5-min duration. Like BUFAR, NRFAR computes statistical features of JM-events to identify foraging activities. NRFAR uses the CBEBA method to recognize JM-events into four classes: \emph{rumination-chews}, \emph{grazing-chews}, \emph{bites}, and \emph{chew-bites}. The NRFAR method represents a significant improvement over the previous BUFAR method, which only distinguished between \emph{bites}, \emph{chew-bites}, and \emph{chews}, without discriminating between \emph{rumination-chews} and \emph{grazing-chews} events. 
The JMFAR method uses a different approach that does not require the identification of JM-events to delimit grazing and rumination bouts. Instead, it extracts information from the detected JM in the segment.

The results showed that the average correct recognition rate of the activities of interest (\emph{grazing} and \emph{rumination}) for NRFAR was 91.5\% \textcolor{black}{when evaluating in DS1}, exceeding BUFAR by 12.0\% and JMFAR by 7.2\% (Figure~\ref{fig_5}). Importantly, this improvement in activity recognition was achieved without incurring substantial changes in computational cost. The remarkable performance improvement of NRFAR was due to the improved discrimination of JM-events produced during rumination and grazing by CBEBA. 
The good classification rate of JM-events allowed the computation of a confidence set of activity features with more specific discriminatory information than BUFAR and JMFAR to enhance activity classifications.
NRFAR presented a minimal confusion of $\leq1.2\%$ between \emph{grazing} and \emph{rumination}, which was lower than the confusion reported by BUFAR ($\geq11.2\%$) and JMFAR ($\geq5.1\%$). The authors hypothesized that the misclassification of foraging activities was reduced because it depends mainly on the misrecognition of JM-events associated with rumination (\emph{rumination-chew}) and grazing (\emph{grazing-chew}, \emph{bite}, and \emph{chew-bite}), and not between all possible JM-event classes. Therefore, NRFAR was less sensitive to JM-events misclassification than BUFAR. Likewise, discrimination between foraging activities and other activities presented a greater error in the NRFAR ($\geq4.1\%$). This confusion was also observed in BUFAR and JMFAR and could be related to the great diversity of behavior represented by the \emph{other} class. From a productivity standpoint, confusion of 5\% or more between \emph{grazing} and \emph{rumination} can significantly affect the diagnoses of feeding performance (e.g. low dry matter intake)~\citep{Watt2015-ek} or metabolic imbalances of nutritional origin in ruminants (e.g., subacute ruminal acidosis)~\citep{Beauchemin2018-dy}.

\textcolor{black}{An acoustic method must be able to work effectively in different setups to have practical utility.
NRFAR, JMFAR, and BUFAR, initially trained using DS1 signals, were tested on DS2 signals. 
Again, NRFAR exceeded the average recognition rate of \emph{grazing} and \emph{rumination} of JMFAR and BUFAR by 4.0\% and 24.7\%, respectively, with higher average balanced accuracy (87.4\% for NRFAR, 84.4\% for JMFAR, and 73.2\% for BUFAR). 
Moreover, the average balanced accuracy of NRFAR in DS2 was 1.0\% higher than in DS1, with similar recognition rates of the three classes in both datasets (Figure~\ref{fig_5}c and Figure~\ref{fig_5_new}c), demonstrating good generalization capability.
JMFAR also exhibited good generalization performance (average balanced accuracy of 78.9\% in DS1 and 84.4\% in DS2) but an improvement in the recognition of \emph{rumination} was compensated with a decrease in \emph{grazing} (Figure~\ref{fig_5}b and Figure~\ref{fig_5_new}b).
Noteworthy was the limited generalization ability of BUFAR to identify \emph{grazing}, decreasing from 83.5\% in DS1 to 57.4\% in DS2 (Figure~\ref{fig_5}a and Figure~\ref{fig_5_new}a).}

Acoustic methods often have lower performance in confined environments such as barns because of the high levels and varying types of noise present there. Acoustic reverberation existing in confined environments is the cause that noise has to be considered convolutional. In free-ranging conditions, noise is still present but is less intense and frequent, and can be considered additive. To reduce the unwanted effects of acoustic noise, an appropriate microphone setup (as shown in Figure~\ref{fig_3}) can be used. Hence, the proper operation of acoustic methods in free-ranging is not necessarily compromised. The effectiveness of an acoustic foraging activity recognizer depends on its ability to work well in adverse field conditions, making it a useful and effective tool for farmers and handlers. In this study, the noise robustness of NRFAR was evaluated and compared with previous methods by adding artificial noises to the original audio signals \textcolor{black}{of DS1 }at different levels ($20~\leq~$SNR$~\leq~-15$~dB), which were even higher than those produced by real noises in classical pasture environments~\citep{bishop2019livestock}. The noise robustness of the methods using a stationary noise source with different properties was evaluated (Figure~\ref{fig_6}). Artificial random Gaussian white noise was used to contaminate the audio signals. The white noise signal has a theoretical \emph{``infinite''} bandwidth and a constant power spectral density across all frequencies, which can degrade important acoustic cues over the entire frequency range. 
NRFAR had great robustness to noise for SNR~$\geq$~10~dB, keeping their balanced accuracy almost constant.
However, the performances of the JMFAR and BUFAR methods decreased with decreasing SNR.
JMFAR performed better than BUFAR at low levels of noise (SNR~$\geq$~10~dB) since the noise had a similar impact on both methods in this SNR range.
BUFAR outperformed JMFAR for moderate and high noise levels (SNR$~\leq~-5$~dB) due to the higher robustness to noise of the JM information from recognized JM-events used by BUFAR.
Furthermore, JMFAR exhibited the largest drop in performance in this experiment.
The decreasing performance of JMFAR was due to the limited robustness to noise of the JM information, computed from detected JM-events, analyzed to recognize foraging activities (Figure~\ref{fig_4}).
Additionally, NRFAR outperformed the other methods for the entire range considered in these numerical experiments (SNR~$\geq$~-15~dB) (14 of 16 evaluated scenarios).

The effects of different nonstationary noise sources commonly present on pastures, such as sounds produced by animals, vehicles, weather, and a mixture of these sounds, were also evaluated. Figure~\ref{fig_7} showed that JMFAR outperformed BUFAR, which is consistent with the results of~\citet{CHELOTTI202369}.
In addition, NRFAR outperformed the previous methods in~61 of~64 evaluated scenarios, with~39 of those cases showing statistical significance (p~<~0.05), as in the evaluations using Gaussian white noise (Figure~\ref{fig_6}).
It should be noted that the largest differences in favor of NRFAR were observed for SNR~$\geq$~15~dB and SNR~$\leq$~0~dB, but NRFAR performed similarly to JMFAR for 10~$\leq$~SNR~$\leq$~5~dB.
Under high noise conditions, the performance of NRFAR was due to the high noise robustness and discriminative power of the JM features used to classify the JM-events by CBEBA (middle level of Figure~\ref{fig_1})~\citep{Martinez-Rau2022-ya}.

The robustness of each method to different noise sources was analyzed. 
The performance of NRFAR using the four nonstationary noise sources was similar to each other for a particular SNR level (Figure~\ref{fig_8}.a), even though these noise sources have different spectral energy distributions~\citep{ozmen2022sound}. 
A similar situation was observed for JMFAR (Figure~\ref{fig_8}b), but not for BUFAR (Figure~\ref{fig_8}c). 
It was noteworthy that NRFAR performed better when evaluated with stationary Gaussian white noise compared to the nonstationary noise sources (Figure~\ref{fig_8}a), particularly for moderate and high noise conditions. This particular situation was also observed in BUFAR (Figure~\ref{fig_8}c).
nonstationary noise sources have uncertain onset, offset, and duration, which can lead to false detection of JM, classifying noises as JM-events (middle level of Figure~\ref{fig_1}).
Figure~\ref{fig_8}b showed that JMFAR performed similarly with all nonstationary noise sources for SNR~$\geq$~-5~dB because it did not depend on the identification of JM-events.
Remarkably, JMFAR was less robust to stationary Gaussian white noise than to stationary noise sources at low noise levels (SNR~$\geq$~5~dB).

NRFAR has a low computational cost of 43,060~ops/s, which is of the same order of magnitude as BUFAR and JMFAR.
It is important to note that most of the computational cost required by NRFAR (43,121~ops/s) comes from the computation of CBEBA (43,118~ops/s) (see~\ref{Appendix_A}). 
This suggests that NRFAR could potentially be implemented in an application-specific ultra-low-power microprocessor, similar to the implementation of CBEBA~\citep{Martinez-Rau2022-processor}.
This computational cost value is theoretical and considers only the arithmetic and logic operations required to execute NRFAR. It is useful to compare the computational requirements of different methods independently on the platform.
However, the total processing time of a constrained electronic device depends on available hardware resources~\citep{manor2022}.
The recent deployment of NRFAR in a low-power microcontroller~\citep{Martinez_SAS2023}, combined with its strong noise robustness, positions NRFAR as a reliable tool to be embedded in an acoustic sensor for recognizing grazing and rumination activities.

\section{Conclusion}
This study proposes an improvement over former acoustic methods to recognize and delimit foraging activity bouts of grazing cattle. Inspired by the former BUFAR method, the proposed NRFAR method analyzes fixed-length segments of recognized JM-events. NRFAR uses a robust JM recognizer that discriminates JM-events produced during grazing and rumination under different operating conditions. 
This allows NRFAR to recognize foraging activities in free-range scenarios, even under adverse acoustic conditions.
The method has shown a significant performance improvement over state-of-the-art acoustic methods in quiet and noisy conditions\textcolor{black}{, and in different settings}. The evaluation of noise robustness was performed by adding artificially different amounts of stationary Gaussian white noise, and nonstationary natural noise commonly present in free-range. Future work must include changes in the analysis of fixed-length segments to variable-length segments using dynamic segmentation to facilitate more accurate estimation of the foraging bouts of interest. Likewise, NRFAR could be used as a reference for developing new methods based on multi-modal data sensors to recognize feeding activities in more adverse environments, such as barns.

\section*{Acknowledgment}
The authors wish to express their gratitude to the staff of the KBS Robotic Dairy Farm, who participated in the investigation. Additionally, we acknowledge the direct support from AgBioResearch-MSU. The authors would like to thank Constanza Quaglia (technical staff, CONICET) and J. Tom\'as Molas G. (technical staff, UNER-UNL) for their technical support in achieving the web demo.
This work was supported by the Universidad Nacional del Litoral [CAID 50620190100080LI and 50620190100151LI]; Universidad Nacional de Rosario [AGR216, 2013 - AGR266, 2016 - and 80020180300053UR, 2019]; Agencia Santafesina de Ciencia, Tecnolog\'ia e Innovaci\'on [IO-2018–00082], CONICET [PUE sinc(I), 2017]; and USDA-NIFA [MICL0222 and MICL0406].

\section*{CRediT authorship contribution statement}
LSMR, JOC, MF, HLR, and LLG participated in conceptualization;
LSMR participated in software stage;
LSMR, JOC, JRG, and AMP participated in the data curation;
LSMR, JOC, MF, HLR and LLG participated in the formal analysis;
LSMR, JOC, MF, and HLR participated in the investigation stage;
LSMR, JOC, MF, HLR, and LLG participated in methodology, validation and visualization stages;
JRG, LLG, SAU, and HLR participated in the funding acquisition;
JRG, LLG, and HLR participated in project administration;
LSMR, JOC, MF, JRG, SAU, AMP, HLR and LLG contributed to the writing and reviewing of the original draft;
All the authors reviewed and approved the manuscript.

\section*{Data availability}
Data will be available on request.

\section*{Declaration of competing interest}
The authors declare that they have no known competing financial interests or personal relationships that could have appeared to influence the work reported in this paper.

\appendix
\section{Computational cost}
\label{Appendix_A}
The computational cost of NRFAR depends on the input audio sampling frequency, the sub-sampling frequency used internally in CBEBA (fixed at $f_{s}=150~Hz$ in this analysis, according to its optimal value), the configuration of the two MLP neural networks used to classify the JM-events and foraging activities, and the duration of the segment lengths (fixed at 5~min). To obtain a valid comparison with other methods, an input sampling frequency of $f_{i}=2~kHz$ and 2~JM-events per second was chosen. Furthermore, the worst-case computational cost scenario was selected for both MLP classifiers. In addition, any arithmetic operation, arithmetic shift, logic comparison, or activation function is counted as one operation. The required number of operations per second for the computation stages of each level of NRFAR is:

\emph{Bottom level}:
\begin{enumerate}
    \item 
    Audio pre-processing: limiting the bandwidth with a second-order band-pass filter and computing the instantaneous power signal requires $7*f_{i}$ and $f_{i}$~ops/s per sample, respectively. Then, 16,000~ops/s are required.
    \item 
    Signal computation: computing and decimating the envelope signal requires $11*f_{i}+150$~ops/s. Computing the energy signal by frames requires $f_{i}+300$~ops/s. Altogether, this stage requires 24,450~ops/s.
\end{enumerate}

\emph{Middle level}:
\begin{enumerate}
    \item 
    JM-event detection: $4+0.925*f_{s}$ and $12+f_{s}$ operations per JM-event are necessary to detect and delimit the boundaries of JM-events. Then, this stage takes 610~ops/s.
    \item 
    Feature extraction: $3.5*f_{s}$ operations per JM-event are necessary to compute the set of JM features. In total, 1050~ops/s are required.
    \item 
    JM-event classification: deciding whether an event should be classified requires $f_{s}+3$ operations per JM event, whereas the MLP with 5-6-4~neurons requires 131~operations per JM-event, thus, 568~ops/s are required.
    \item 
    Tuning parameters: $f_{s}+39$ operations per JM-event are necessary to update the thresholds. Then, 378~ops/s are required.
\end{enumerate}

\emph{Middle level}:
\begin{enumerate}
    \item 
    Segment buffering: this stage requires 2~operations per JM-event equivalent to 4~ops/s.
    \item 
    Feature extraction: computing the set of activity features requires 608~ops/segment.
    \item 
    Activity classification: considering the maximum number of neurons~(10) in the hidden layer, the MLP requires 185~ops/segment.
    \item 
    Smoothing process: this filtering stage takes 2~ops/segment.
\end{enumerate}

Finally, the total computational cost of NRFAR is 43,060~ops/s~+~795~ops/segment $\approx$ 43,063 ops/s. Similar to BUFAR, the overall computational cost almost exclusively depends on the bottom and middle levels of Figure~\ref{fig_1} (i.e., the JM event recognizer) because the top level is only executed once every 5~min (segment length). Hence, the total computational cost of NRFAR can be expressed as 12,918,795~ops/segment.

\section{Statistical hypothesis test}
\label{Appendix_B}

The statistically significant discrepancies in the balanced accuracy between NRFAR and BUFAR, NRFAR and JMFAR, and JMFAR and BUFAR were evaluated using the Wilcoxon signed-rank test~\citep{Wilcoxon1945-zq}. Tables~\ref{table_b1}, \ref{table_b2}, and~\ref{table_b3} show the p-values obtained from the comparison of these methods. P-values with a green background indicate a significant difference in performance with a confidence level of 5\% (p~=~0.05), and p-values with a pink background indicate a nonsignificant difference.

%TABLA B1
\begin{table}[!htb]
\caption{Statistically significant p-values were obtained by comparing the performance of the NRFAR and BUFAR methods with different noise sources at several noise levels.}
\label{table_b1}
\centering
\begin{tabular}{|c|ccccc|}
\hline
\multicolumn{1}{|c|}{} & \multicolumn{5}{c|}{NRFAR vs BUFAR} \\ \cline{2-6} 
\multicolumn{1}{|c|}{\multirow{-2}{*}{SNR {[}dB{]}}} & Animals & Vehicles & Weather & Mixture & White \\ \hline
20 & \cellcolor[HTML]{B6D7A8}3.88e-05 & \cellcolor[HTML]{B6D7A8}1.69e-08 & \cellcolor[HTML]{B6D7A8}8.75e-06 & \cellcolor[HTML]{B6D7A8}5.36e-06 & \cellcolor[HTML]{B6D7A8}1.02e-08 \\
15 & \cellcolor[HTML]{B6D7A8}1.21e-04 & \cellcolor[HTML]{B6D7A8}7.79e-04 & \cellcolor[HTML]{B6D7A8}5.38e-04 & \cellcolor[HTML]{B6D7A8}8.33e-04 & \cellcolor[HTML]{B6D7A8}3.30e-11 \\
10 & \cellcolor[HTML]{B6D7A8}1.58e-10 & \cellcolor[HTML]{B6D7A8}3.78e-01 & \cellcolor[HTML]{B6D7A8}9.34e-04 & \cellcolor[HTML]{B6D7A8}1.93e-06 & \cellcolor[HTML]{B6D7A8}7.36e-14 \\
5 & \cellcolor[HTML]{B6D7A8}1.04e-15 & \cellcolor[HTML]{B6D7A8}1.92e-06 & \cellcolor[HTML]{B6D7A8}9.88e-15 & \cellcolor[HTML]{B6D7A8}1.34e-15 & \cellcolor[HTML]{B6D7A8}4.36e-13 \\
0 & \cellcolor[HTML]{B6D7A8}1.43e-09 & \cellcolor[HTML]{B6D7A8}1.57e-09 & \cellcolor[HTML]{B6D7A8}1.71e-15 & \cellcolor[HTML]{B6D7A8}4.59e-10 & \cellcolor[HTML]{B6D7A8}1.16e-05 \\
-5 & \cellcolor[HTML]{B6D7A8}7.39e-04 & \cellcolor[HTML]{B6D7A8}8.82e-06 & \cellcolor[HTML]{B6D7A8}5.20e-05 & \cellcolor[HTML]{B6D7A8}6.53e-04 & \cellcolor[HTML]{E6B8AF}1.98e-01 \\
-10 & \cellcolor[HTML]{E6B8AF}6.23e-01 & \cellcolor[HTML]{B6D7A8}1.19e-02 & \cellcolor[HTML]{E6B8AF}9.68e-01 & \cellcolor[HTML]{E6B8AF}9.04e-01 & \cellcolor[HTML]{E6B8AF}2.16e-01 \\
-15 & \cellcolor[HTML]{E6B8AF}5.63e-01 & \cellcolor[HTML]{E6B8AF}1.85e-01 & \cellcolor[HTML]{E6B8AF}9.44e-01 & \cellcolor[HTML]{E6B8AF}4.19e-01 & \cellcolor[HTML]{B6D7A8}6.01e-04 \\ \hline
\end{tabular}
\end{table}

%TABLA B2
\begin{table}[!htb]
\caption{Statistically significant p-values were obtained by comparing the performance of the NRFAR and JMFAR methods with different noise sources at several noise levels.}
\label{table_b2}
\centering
\begin{tabular}{|c|ccccc|}
\hline
\multicolumn{1}{|c|}{} & \multicolumn{5}{c|}{NRFAR vs JMFAR} \\ \cline{2-6} 
\multicolumn{1}{|c|}{\multirow{-2}{*}{SNR {[}dB{]}}} & Animals & Vehicles & Weather & Mixture & White \\ \hline
20 & \cellcolor[HTML]{E6B8AF}8.45e-02 & \cellcolor[HTML]{B6D7A8}6.52e-04 & \cellcolor[HTML]{B6D7A8}1.80e-03 & \cellcolor[HTML]{B6D7A8}6.95e-03 & \cellcolor[HTML]{B6D7A8}5.45e-05 \\
15 & \cellcolor[HTML]{E6B8AF}5.55e-01 & \cellcolor[HTML]{E6B8AF}2.30e-01 & \cellcolor[HTML]{E6B8AF}1.61e-01 & \cellcolor[HTML]{E6B8AF}9.76e-01 & \cellcolor[HTML]{B6D7A8}6.11e-10 \\
10 & \cellcolor[HTML]{E6B8AF}3.66e-01 & \cellcolor[HTML]{E6B8AF}7.02e-01 & \cellcolor[HTML]{E6B8AF}3.28e-01 & \cellcolor[HTML]{E6B8AF}9.02e-01 & \cellcolor[HTML]{B6D7A8}2.61e-13 \\
5 & \cellcolor[HTML]{E6B8AF}6.48e-01 & \cellcolor[HTML]{E6B8AF}5.98e-01 & \cellcolor[HTML]{E6B8AF}3.36e-01 & \cellcolor[HTML]{E6B8AF}2.69e-01 & \cellcolor[HTML]{B6D7A8}4.80e-15 \\
0 & \cellcolor[HTML]{B6D7A8}3.12e-02 & \cellcolor[HTML]{B6D7A8}4.20e-04 & \cellcolor[HTML]{B6D7A8}3.77e-02 & \cellcolor[HTML]{E6B8AF}2.14e-01 & \cellcolor[HTML]{B6D7A8}8.13e-20 \\
-5 & \cellcolor[HTML]{B6D7A8}3.29e-06 & \cellcolor[HTML]{B6D7A8}6.08e-07 & \cellcolor[HTML]{B6D7A8}8.82e-03 & \cellcolor[HTML]{B6D7A8}6.31e-03 & \cellcolor[HTML]{B6D7A8}2.83e-13 \\
-10 & \cellcolor[HTML]{B6D7A8}4.04e-02 & \cellcolor[HTML]{B6D7A8}2.96e-03 & \cellcolor[HTML]{B6D7A8}1.20e-02 & \cellcolor[HTML]{B6D7A8}4.94e-03 & \cellcolor[HTML]{B6D7A8}6.17e-08 \\
-15 & \cellcolor[HTML]{E6B8AF}5.95e-01 & \cellcolor[HTML]{E6B8AF}1.71e-01 & \cellcolor[HTML]{E6B8AF}7.00e-01 & \cellcolor[HTML]{E6B8AF}4.54e-01 & \cellcolor[HTML]{B6D7A8}3.15e-09 \\ \hline
\end{tabular}
\end{table}

%TABLA B3
\begin{table}[!htbp]
\caption{Statistically significant p-values were obtained by comparing the performance of the JMFAR and BUFAR methods with different noise sources at several noise levels.}
\label{table_b3}
\centering
\begin{tabular}{|l|ccccc|}
\hline
\multicolumn{1}{|c|}{} & \multicolumn{5}{c|}{JMFAR vs BUFAR} \\ \cline{2-6} 
\multicolumn{1}{|c|}{\multirow{-2}{*}{SNR {[}dB{]}}} & Animals & Vehicles & Weather & Mixture & White \\ \hline
20 & \cellcolor[HTML]{B6D7A8}4.67e-02 & \cellcolor[HTML]{B6D7A8}2.95e-03 & \cellcolor[HTML]{B6D7A8}2.33e-02 & \cellcolor[HTML]{B6D7A8}2.09e-02 & \cellcolor[HTML]{B6D7A8}4.39e-02 \\
15 & \cellcolor[HTML]{B6D7A8}1.79e-04 & \cellcolor[HTML]{B6D7A8}6.66e-03 & \cellcolor[HTML]{B6D7A8}3.74e-03 & \cellcolor[HTML]{B6D7A8}2.36e-03 & \cellcolor[HTML]{E6B8AF}1.73e-01 \\
10 & \cellcolor[HTML]{B6D7A8}2.01e-14 & \cellcolor[HTML]{E6B8AF}7.01e-02 & \cellcolor[HTML]{B6D7A8}4.646e-09 & \cellcolor[HTML]{B6D7A8}1.49e-10 & \cellcolor[HTML]{E6B8AF}1.58e-01 \\
5 & \cellcolor[HTML]{B6D7A8}6.94e-17 & \cellcolor[HTML]{B6D7A8}1.04e-12 & \cellcolor[HTML]{B6D7A8}8.32e-18 & \cellcolor[HTML]{B6D7A8}3.47e-17 & \cellcolor[HTML]{E6B8AF}6.68e-01 \\
0 & \cellcolor[HTML]{B6D7A8}1.25e-06 & \cellcolor[HTML]{B6D7A8}5.57e-10 & \cellcolor[HTML]{B6D7A8}2.58e-11 & \cellcolor[HTML]{B6D7A8}1.50e-10 & \cellcolor[HTML]{B6D7A8}1.07e-14 \\
-5 & \cellcolor[HTML]{E6B8AF}6.81e-02 & \cellcolor[HTML]{E6B8AF}1.38e-01 & \cellcolor[HTML]{E6B8AF}5.61e-01 & \cellcolor[HTML]{E6B8AF}8.14e-01 & \cellcolor[HTML]{B6D7A8}4.71e-16 \\
-10 & \cellcolor[HTML]{B6D7A8}9.58e-09 & \cellcolor[HTML]{B6D7A8}1.53e-04 & \cellcolor[HTML]{B6D7A8}7.81e-06 & \cellcolor[HTML]{B6D7A8}4.03e-08 & \cellcolor[HTML]{B6D7A8}3.89e-09 \\
-15 & \cellcolor[HTML]{B6D7A8}4.20e-04 & \cellcolor[HTML]{E6B8AF}5.00e-01 & \cellcolor[HTML]{B6D7A8}2.73e-02 & \cellcolor[HTML]{B6D7A8}1.05e-04 & \cellcolor[HTML]{B6D7A8}5.31e-06 \\ \hline
\end{tabular}
\end{table}

%% References
%% Usage of~\cite is as follows:
%%  ~\citep{key}          ==>>  [#]
%%  ~\cite[chap. 2]{key} ==>>  [#, chap. 2]
%%  ~\citet{key}         ==>>  Author [#]

%\newpage
% \section*{References}
\clearpage
%\bibliographystyle{apalike}
%\bibliography{bibliography}

\end{document}


\section*{Supplementary Material: Waveforms and spectrograms of audio signals}

\begin{figure}[!h]
    \centering
    \includegraphics[width=0.95\textwidth]{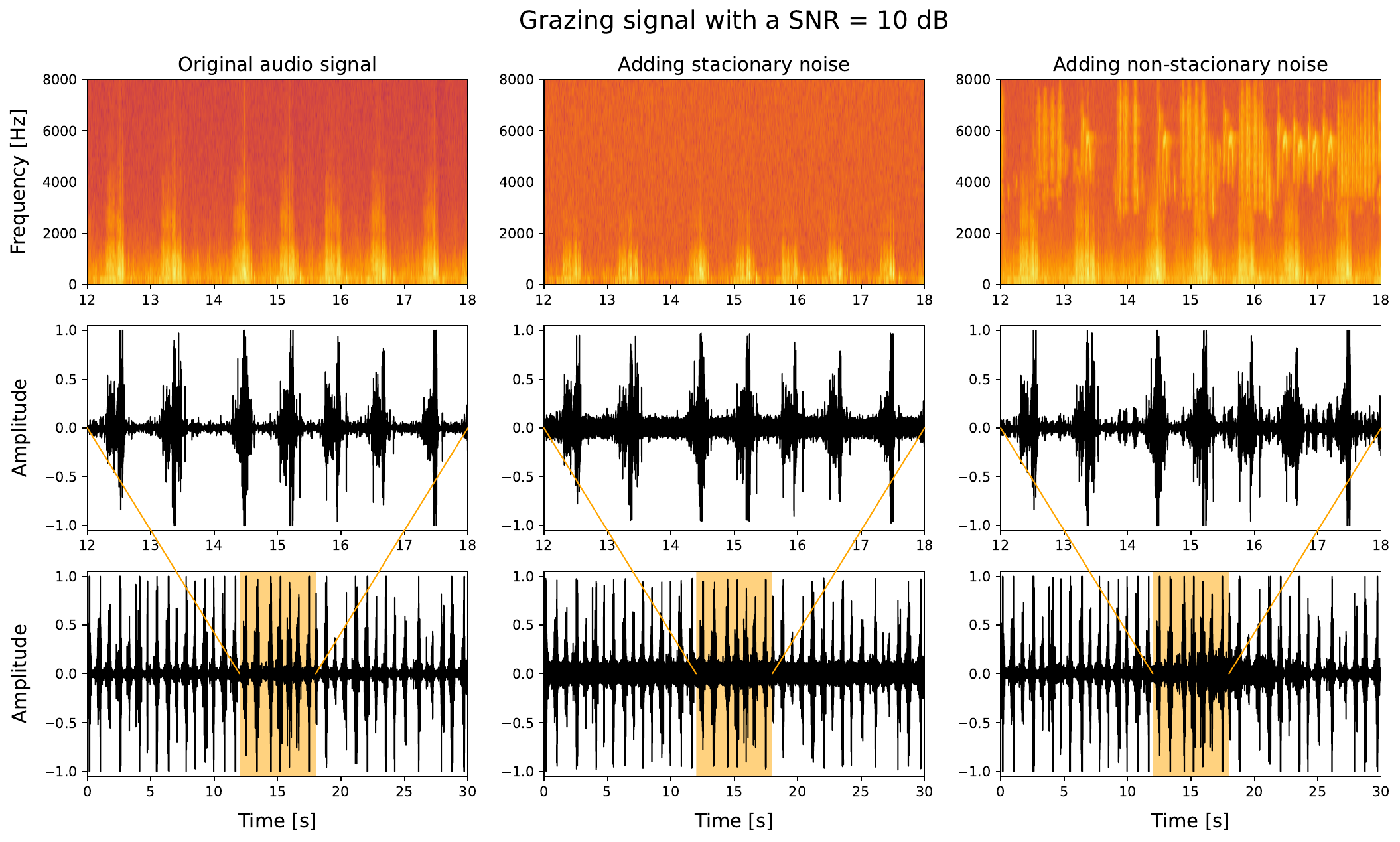}
    \caption{Waveform and spectrogram of an audio signal in grazing condition contaminated with additive noise achieving a signal-to-noise ratio (SNR) of 10 dB. The original audio signal (left panels) is contaminated with Gaussian white noise (middle panels) or sounds present in the pasture (right panels).}
    \label{fig_s1}
\end{figure}

\begin{figure}[!h]
    \centering
    \includegraphics[width=0.95\textwidth]{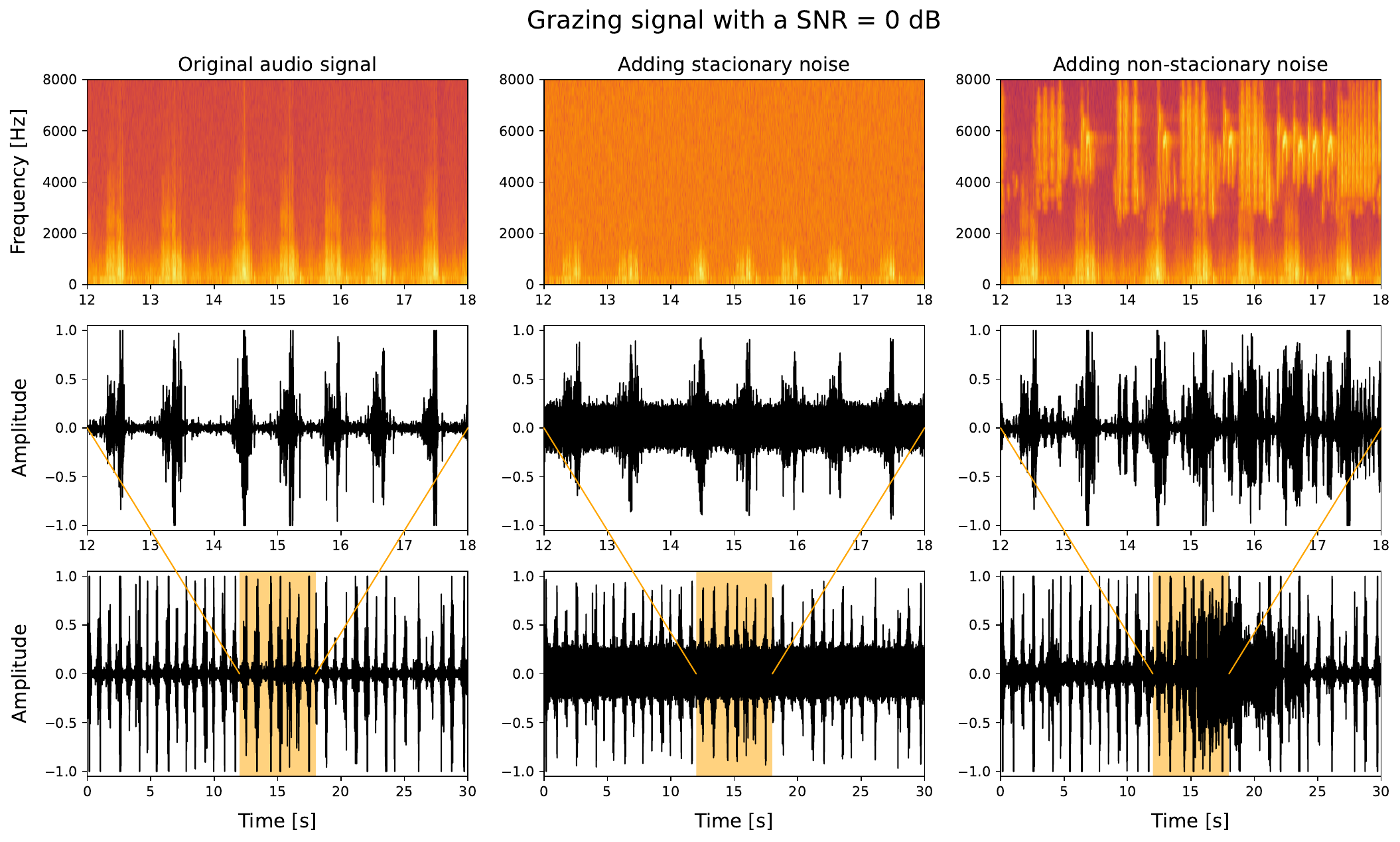}
    \caption{Waveform and spectrogram of an audio signal in grazing condition contaminated with additive noise achieving a signal-to-noise ratio (SNR) of 0 dB. The original audio signal (left panels) is contaminated with Gaussian white noise (middle panels) or sounds present in the pasture (right panels).}
    \label{fig_s2}
\end{figure}

\begin{figure}[!h]
    \centering
    \includegraphics[width=0.95\textwidth]{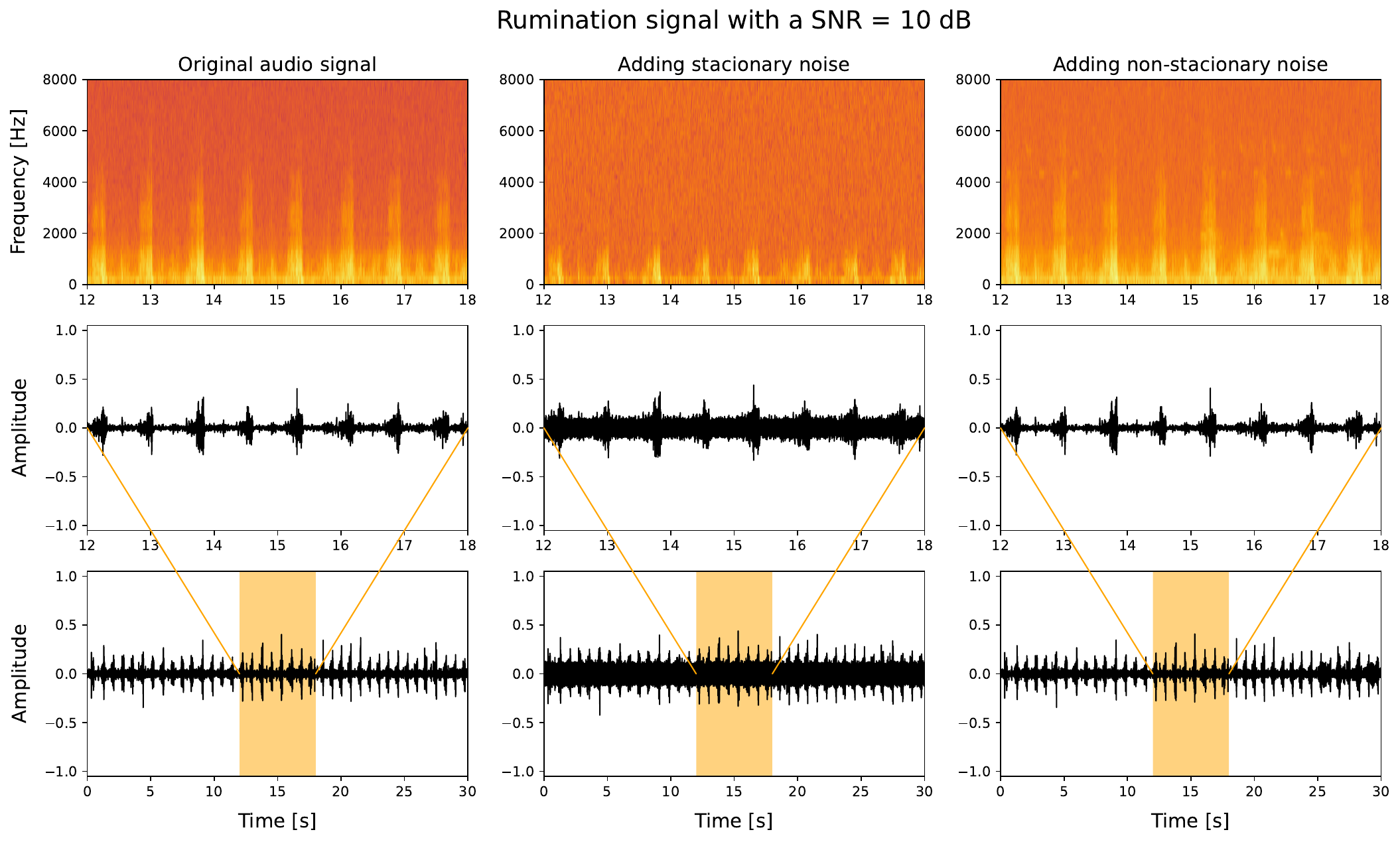}
    \caption{Waveform and spectrogram of an audio signal in rumination condition contaminated with additive noise achieving a signal-to-noise ratio (SNR) of 10 dB. The original audio signal (left panels) is contaminated with Gaussian white noise (middle panels) or sounds present in the pasture (right panels).}
    \label{fig_s3}
\end{figure}

\begin{figure}[!h]
    \centering
    \includegraphics[width=0.95\textwidth]{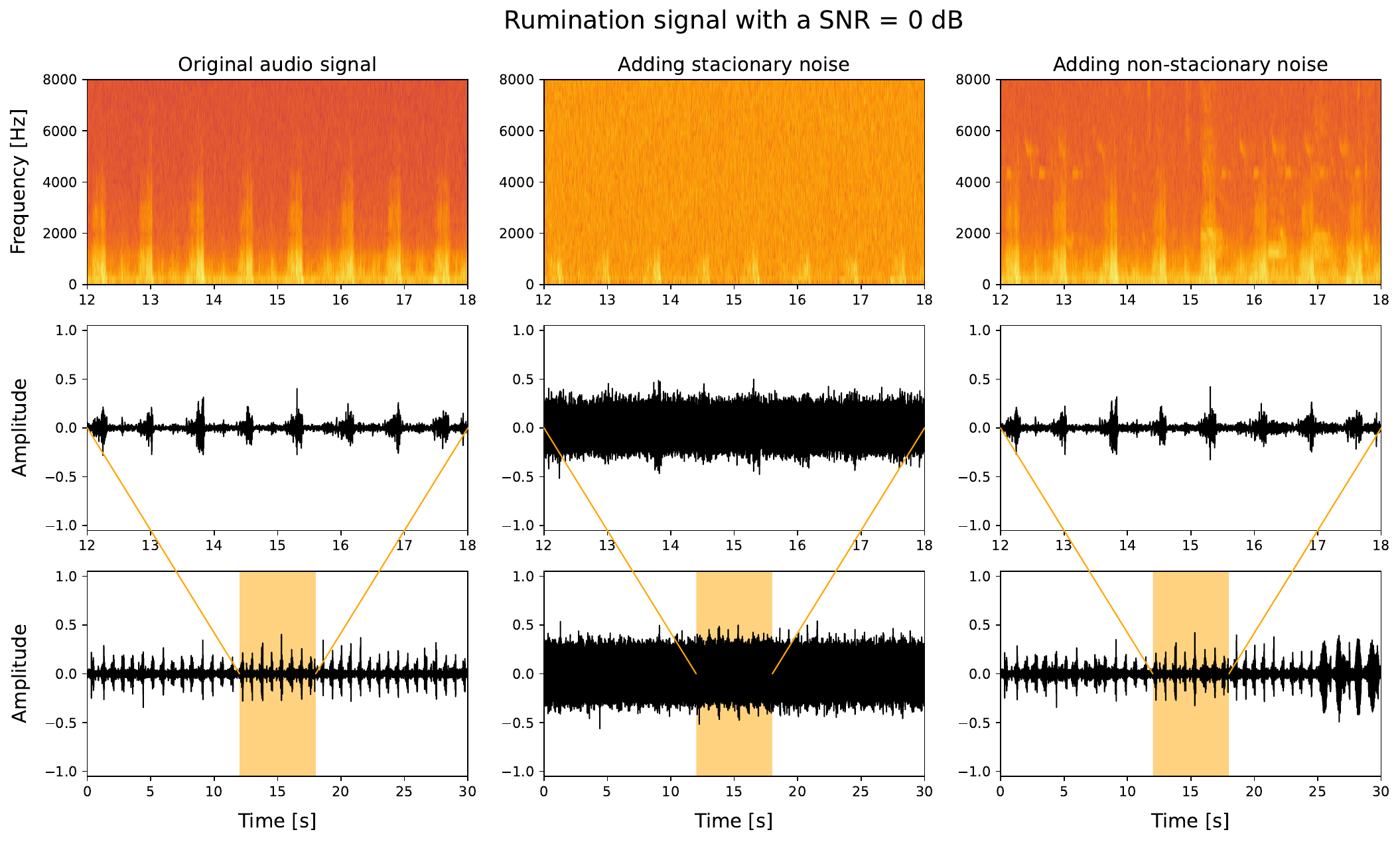}
    \caption{Waveform and spectrogram of an audio signal in rumination condition contaminated with additive noise achieving a signal-to-noise ratio (SNR) of 0 dB. The original audio signal (left panels) is contaminated with Gaussian white noise (middle panels) or sounds present in the pasture (right panels).}
    \label{fig_s4}
\end{figure}

Fig.~\ref{fig_s1} to Fig.~\ref{fig_s4} show waveforms and spectrograms of fragments of audio signals used in this work. Signals contaminated with additive noise are not normalized to obtain a better graphical representation.

%FIG 1

%FIG 2

%FIG 3

%FIG 4
